\documentclass[10pt,twocolumn,letterpaper]{article}

\usepackage{mystyle}

\cvprfinalcopy 

\def\cvprPaperID{10659} 
\def\confYear{CVPR 2021}

\ifcvprfinal\fi
\begin{document}

\makeatletter
\newcommand{\printfnsymbol}[1]{%
  \textsuperscript{\@fnsymbol{#1}}%
}
\makeatother

\title{Natural Adversarial Examples}

\author{Dan Hendrycks\\
UC Berkeley\\
\and
Kevin Zhao\thanks{Equal Contribution.}\\
University of Washington\\
\and
Steven Basart\printfnsymbol{1}\\
UChicago\\
\and
Jacob Steinhardt, Dawn Song\\
UC Berkeley\\
}

\maketitle

\begin{abstract}
We introduce two challenging datasets that reliably cause machine learning model performance to substantially degrade.
The datasets are collected with a simple adversarial filtration technique to create datasets with limited spurious cues.
Our datasets' real-world, unmodified examples transfer to various unseen models reliably, demonstrating that computer vision models have shared weaknesses.
The first dataset is called \textsc{ImageNet-A} and is like the ImageNet test set, but it is far more challenging for existing models. We also curate an adversarial out-of-distribution detection dataset called \textsc{ImageNet-O}, which is the first out-of-distribution detection dataset created for ImageNet models. On \textsc{ImageNet-A} a DenseNet-121 obtains around 2\% accuracy, an accuracy drop of approximately 90\%, and its out-of-distribution detection performance on \textsc{ImageNet-O} is near random chance levels. We find that existing data augmentation techniques hardly boost performance, and using other public training datasets provides improvements that are limited. However, we find that improvements to computer vision architectures provide a promising path towards robust models.
\end{abstract}

\section{Introduction}

\begin{figure}[t]
    \centering
    \includegraphics[width=0.47\textwidth]{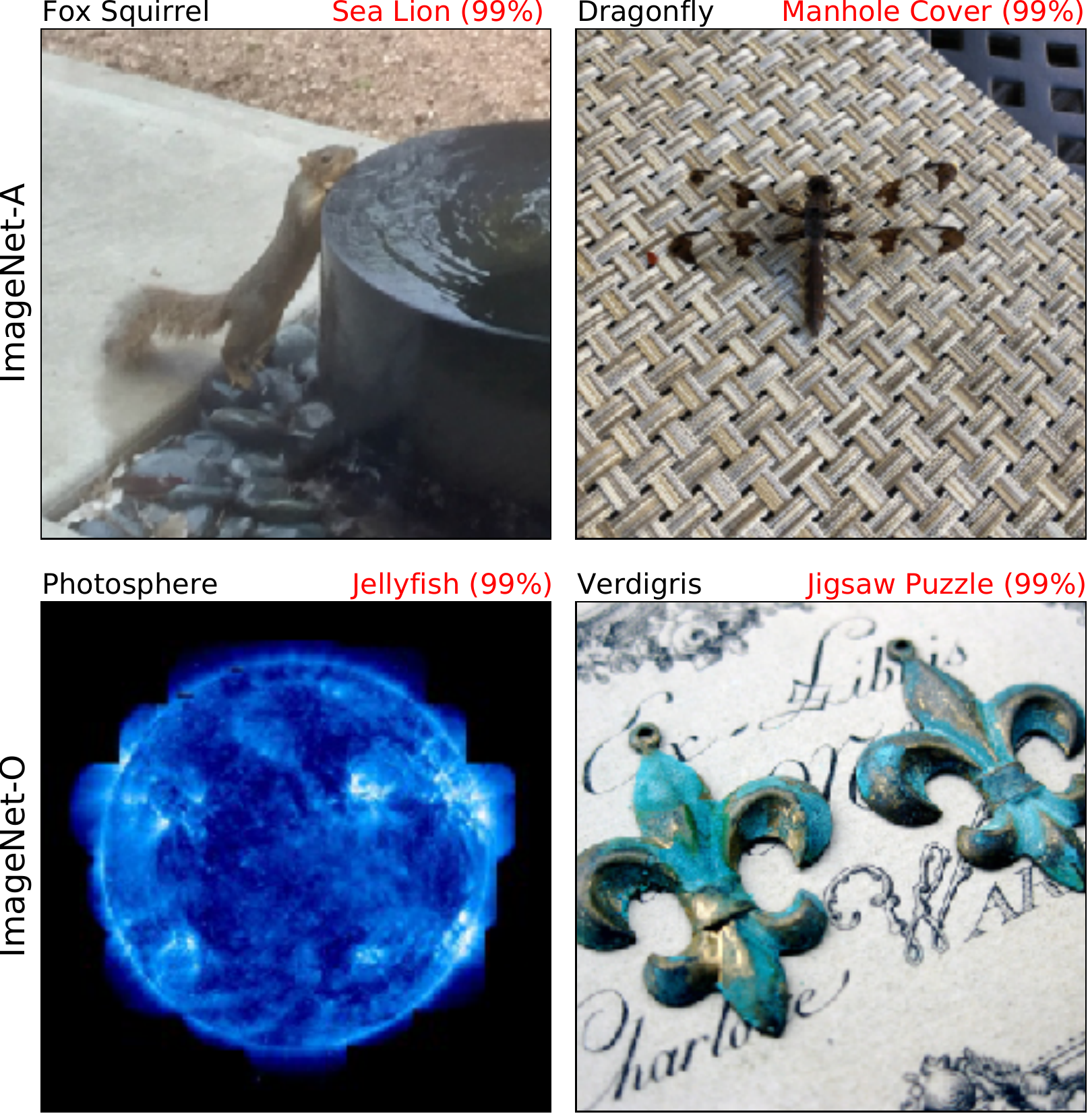}
\caption{Natural adversarial examples from \textsc{ImageNet-A} and \textsc{ImageNet-O}. The black text is the actual class, and the red text is a ResNet-50 prediction and its confidence. \textsc{ImageNet-A} contains images that classifiers should be able to classify, while \textsc{ImageNet-O} contains anomalies of unforeseen classes which should result in low-confidence predictions. ImageNet-1K models do not train on examples from ``Photosphere'' nor ``Verdigris'' classes, so these images are anomalous. Most natural adversarial examples lead to wrong predictions despite occurring naturally.
}\label{fig:splash}
\vspace{-10pt}
\end{figure}

Research on the ImageNet \cite{imagenet} benchmark has led to numerous advances in classification \cite{AlexNet}, object detection \cite{Huang2017SpeedAccuracyTF}, and segmentation \cite{He2018MaskR}. 
ImageNet classification improvements are broadly applicable and highly predictive of improvements on many tasks \cite{Kornblith2018DoBI}. Improvements on ImageNet classification have been so great that some call ImageNet classifiers ``superhuman'' \cite{He2015DelvingDI}.
However, performance is decidedly subhuman when the test distribution does not match the training distribution \cite{hendrycks2019robustness}. The distribution seen at test-time can include inclement weather conditions and obscured objects, and it can also include objects that are anomalous.\looseness=-1

\begin{figure*}[t]
\vspace{-10pt}
\centering
\begin{subfigure}{.5\textwidth}
\includegraphics[width=0.99\textwidth]{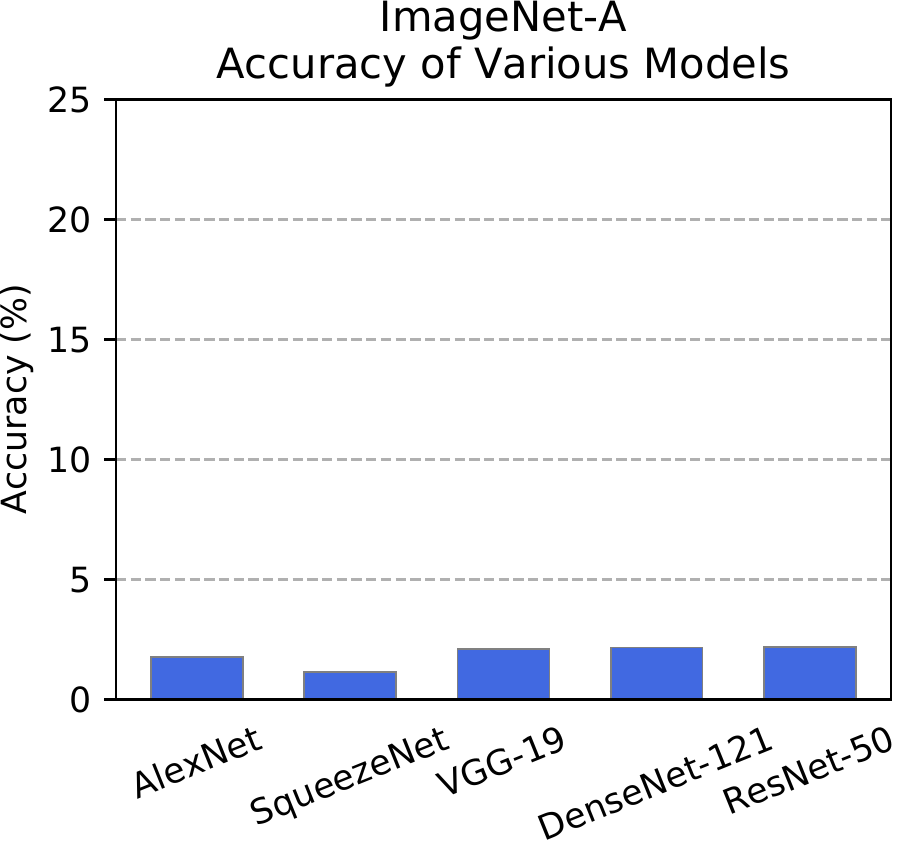}
\end{subfigure}%
\begin{subfigure}{.5\textwidth}
\includegraphics[width=0.99\textwidth]{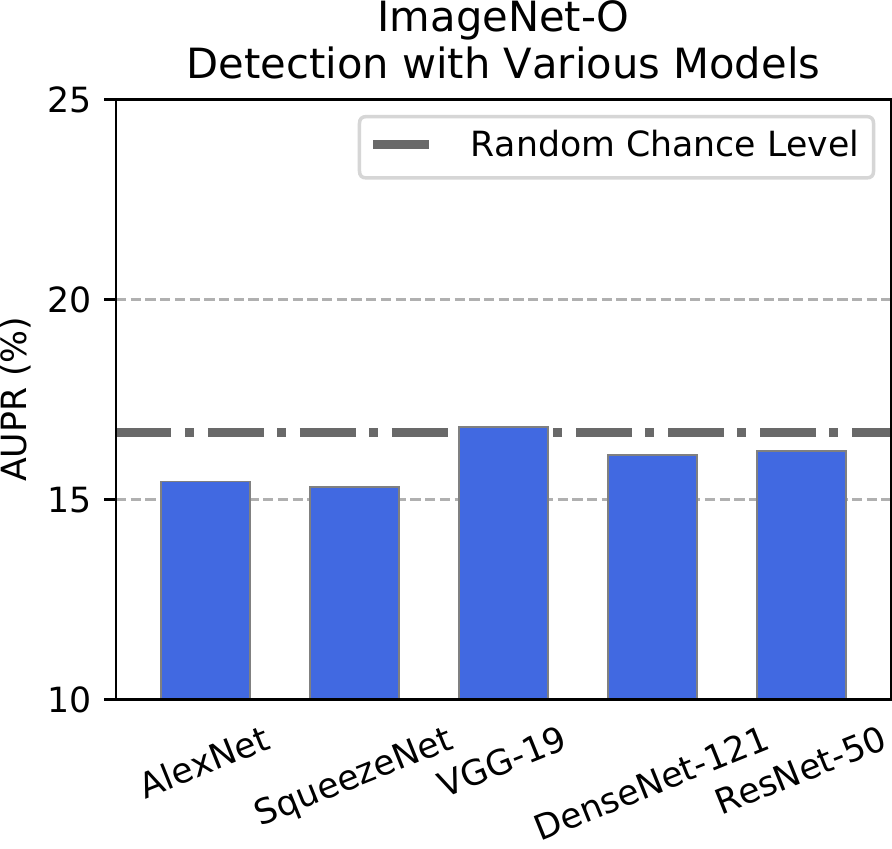}
\end{subfigure}
\caption{
Various ImageNet classifiers of different architectures fail to generalize well to \textsc{ImageNet-A} and \textsc{ImageNet-O}.
Higher Accuracy and higher AUPR is better. See \Cref{sec:experiments} for a description of the AUPR out-of-distribution detection measure. These specific models were not used in the creation of \textsc{ImageNet-A} and \textsc{ImageNet-O}, so our adversarially filtered image transfer across models. 
\looseness=-1
}\label{fig:modelsono}
\vspace{-10pt}
\end{figure*}

Recht et al., 2019 \cite{Recht2019DoIC} remind us that ImageNet test examples tend to be simple, clear, close-up images, so that the current test set may be too easy and may not represent harder images encountered in the real world. Geirhos et al., 2020 argue that image classification datasets contain ``spurious cues'' or ``shortcuts'' \cite{Geirhos2020ShortcutLI, Arjovsky2019InvariantRM}. For instance, models may use an image's background to predict the foreground object's class; a cow tends to co-occur with a green pasture, and even though the background is inessential to the object's identity, models may predict ``cow'' primarily using the green pasture background cue. When datasets contain spurious cues, they can lead to performance estimates that are optimistic and inaccurate.\looseness=-1

To counteract this, we curate two hard ImageNet test sets of natural adversarial examples with adversarial filtration. By using adversarial filtration, we can test how well models perform when simple-to-classify examples are removed, which includes examples that are solved with simple spurious cues. Some examples are depicted in \Cref{fig:splash}, which are simple for humans but hard for models. Our examples demonstrate that it is possible to reliably fool many models with clean natural images, while previous attempts at exposing and measuring model fragility rely on synthetic distribution corruptions \cite{geirhos,hendrycks2019robustness}, artistic renditions \cite{Hendrycks2020TheMF}, and adversarial distortions.\looseness=-1

We demonstrate that clean examples can reliably degrade and transfer to other unseen classifiers using our first dataset. We call this dataset \textsc{ImageNet-A}, which contains images from a distribution unlike the ImageNet training distribution. \textsc{ImageNet-A} examples belong to ImageNet classes, but the examples are harder and can cause mistakes across various models. They cause consistent classification mistakes due to scene complications encountered in the long tail of scene configurations and by exploiting classifier blind spots (see \Cref{sec:failures}). Since examples transfer reliably, this dataset shows models have unappreciated shared weaknesses.

The second dataset allows us to test model uncertainty estimates when semantic factors of the data distribution shift. Our second dataset is \textsc{ImageNet-O}, which contains image concepts from outside ImageNet-1K. These out-of-distribution images reliably cause models to mistake the examples as high-confidence in-distribution examples. To our knowledge this is the first dataset of anomalies or out-of-distribution examples developed to test ImageNet models. While \textsc{ImageNet-A} enables us to test image classification performance when the \emph{input data distribution shifts}, \textsc{ImageNet-O} enables us to test out-of-distribution detection performance when the \emph{label distribution shifts}.

We examine methods to improve performance on adversarially filtered examples. However, this is difficult because \Cref{fig:modelsono} shows that examples successfully transfer to unseen or black-box models.
To improve robustness, numerous techniques have been proposed. We find data augmentation techniques such as adversarial training decrease performance, while others can help by a few percent. We also find that a $10\times$ increase in training data corresponds to a less than a $10\%$ increase in accuracy. Finally, we show that improving model architectures is a promising avenue toward increasing robustness. Even so, current models have substantial room for improvement. Code and our two datasets are available at \href{https://github.com/hendrycks/natural-adv-examples}{\texttt{github.com/hendrycks/natural-adv-examples}}. 

\section{Related Work}
\noindent\textbf{Adversarial Examples.}\quad Real-world images may be chosen adversarially to cause performance decline. Goodfellow et al. \cite{goodfellowblog} define adversarial examples \cite{adversarial} as ``inputs to machine learning models that an attacker has intentionally designed to cause the model to make a mistake.'' Most adversarial examples research centers around artificial $\ell_p$ adversarial examples, which are examples perturbed by nearly worst-case distortions that are small in an $\ell_p$ sense. Su et al., 2018 \cite{Su2018IsRT} remind us that most $\ell_p$ adversarial examples crafted from one model can only be transferred within the same family of models. However, our adversarially filtered images transfer to all tested model families and move beyond the restrictive $\ell_p$ threat model.

\begin{figure}
\centering
\includegraphics[width=0.48\textwidth]{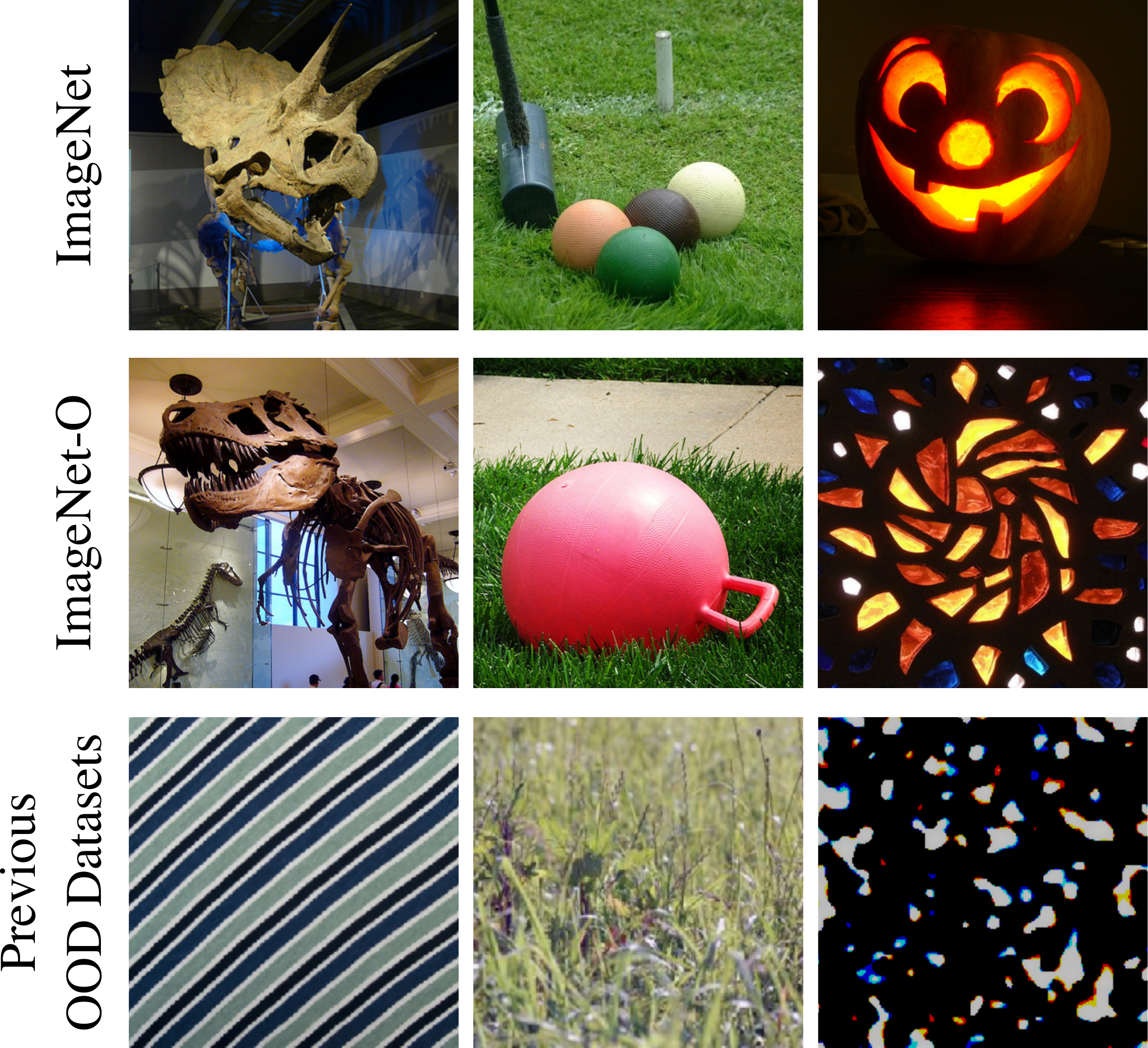}
\caption{
\textsc{ImageNet-O} examples are closer to ImageNet examples than previous out-of-distribution (OOD) detection datasets. For example, ImageNet has triceratops examples and \textsc{ImageNet-O} has visually similar T-Rex examples, but they are still OOD.
Previous OOD detection datasets use OOD examples from wholly different data generating processes. For instance, previous work uses the Describable Textures Dataset \cite{cimpoi14describing}, Places365 scenes \cite{zhou2017places}, and synthetic blobs to test ImageNet OOD detectors.
To our knowledge we propose the first dataset of OOD examples collected for ImageNet models.
}\label{fig:ood}
\vspace{-5pt}
\end{figure}

\noindent\textbf{Out-of-Distribution Detection.}\quad 
For out-of-distribution (OOD) detection \cite{hendrycks17baseline,kimin,hendrycks2019oe,hendrycks2019selfsupervised} models learn a distribution, such as the ImageNet-1K distribution, and are tasked with producing quality anomaly scores that distinguish between usual test set examples and examples from held-out anomalous distributions.
For instance, Hendrycks et al., 2017 \cite{hendrycks17baseline} treat CIFAR-10 as the in-distribution and treat Gaussian noise and the SUN scene dataset \cite{Xiao2010SUNDL} as out-of-distribution data.
They show that the negative of the maximum softmax probability, or the the negative of the classifier prediction probability, is a high-performing anomaly score that can separate in- and out-of-distribution examples, so much so that it remains competitive to this day.
Since that time, other work on out-of-distribution detection has continued to use datasets from other research benchmarks as anomaly stand-ins, producing far-from-distribution anomalies. Using visually dissimilar research datasets as anomaly stand-ins is critiqued in Ahmed et al., 2019 \cite{Ahmed2019DetectingSA}. Some previous OOD detection datasets are depicted in the bottom row of \Cref{fig:ood} \cite{hendrycks2019oe}.
Many of these anomaly sources are unnatural and deviate in numerous ways from the distribution of usual examples. In fact, some of the distributions can be deemed anomalous from local image statistics alone. Next, Meinke et al., 2019 \cite{Meinke2019TowardsNN} propose studying adversarial out-of-distribution detection by detecting adversarially optimized uniform noise. In contrast, we propose a dataset for more realistic adversarial anomaly detection; our dataset contains hard anomalies generated by shifting the distribution's labels and keeping non-semantic factors similar to the original training distribution.

\noindent\textbf{Spurious Cues and Unintended Shortcuts.} Models may learn spurious cues and obtain high accuracy, but for the wrong reasons \cite{Lapuschkin2019UnmaskingCH,Geirhos2020ShortcutLI}. 
Spurious cues are a studied problem in natural language processing \cite{Cai2017PayAT,Gururangan2018AnnotationAI}. Many recently introduced NLP datasets use adversarial filtration to create ``adversarial datasets'' by sieving examples solved with simple spurious cues \cite{Sakaguchi2019WINOGRANDEAA,Bhagavatula2019AbductiveCR,Zellers2019HellaSwagCA,Dua2019DROPAR,Bisk2020PIQARA,Hendrycks2020AligningAW}. Like this recent concurrent research, we also use adversarial filtration \cite{Sung1995LearningAE}, but the technique of adversarial filtration has not been applied to collecting image datasets until this paper. Additionally, adversarial filtration in NLP removes only the easiest examples, while we use filtration to select only the hardest examples and ignore examples of intermediate difficulty.
Adversarially filtered examples for NLP also do \emph{not} reliably transfer even to weaker models. In Bisk et al., 2019 \cite{Bisk2019PIQARA} BERT errors do not reliably transfer to weaker GPT-1 models. This is one reason why it is not obvious \emph{a priori} whether adversarially filtered images should transfer. In this work, we show that adversarial filtration algorithms can find examples that reliably transfer to both weaker and stronger models. Since adversarial filtration can remove examples that are solved by simple spurious cues, models must learn more robust features for our datasets.\looseness=-1 

\noindent\textbf{Robustness to Shifted Input Distributions.}\quad 
Recht et al., 2019 \cite{Recht2019DoIC} create a new ImageNet test set resembling the original test set as closely as possible. 
They found evidence that matching the difficulty of the original test set required selecting images deemed the easiest and most obvious by Mechanical Turkers. However, Engstrom et al., 2020 \cite{Engstrom2020IdentifyingSB} estimate that the accuracy drop from ImageNet to ImageNetV2 is less than $3.6\%$.
In contrast, model accuracy can decrease by over $50\%$ with \textsc{ImageNet-A}. Brendel et al., 2018 \cite{Brendel2018ApproximatingCW} show that classifiers that do not know 
the spatial ordering of image regions can be competitive on the ImageNet test set, possibly due to the dataset's lack of difficulty. Judging classifiers by their performance on easier examples has potentially masked many of their shortcomings. For example, Geirhos et al., 2019 \cite{geirhos2019} artificially overwrite each ImageNet image's textures and conclude that classifiers learn to rely on textural cues and under-utilize information about object shape. Recent work shows that classifiers are highly susceptible to non-adversarial stochastic corruptions \cite{hendrycks2019robustness}. While they distort images with $75$ different algorithmically generated corruptions, our sources of distribution shift tend to be more heterogeneous and varied, and our examples are naturally occurring.

\begin{figure*}[t]
\centering
\begin{subfigure}{.25\textwidth}
    \centering
    \includegraphics[width=0.98\textwidth]{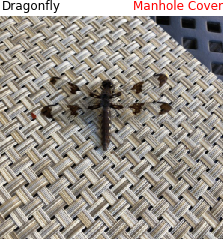}
\end{subfigure}%
\begin{subfigure}{.25\textwidth}
    \centering
    \includegraphics[width=0.98\textwidth]{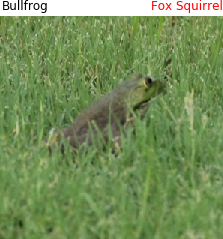}
\end{subfigure}%
\begin{subfigure}{.25\textwidth}
    \centering
    \includegraphics[width=0.98\textwidth]{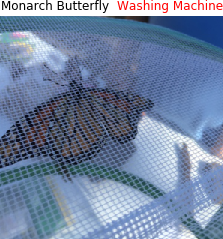}
\end{subfigure}%
\begin{subfigure}{.25\textwidth}
    \centering
    \includegraphics[width=0.98\textwidth]{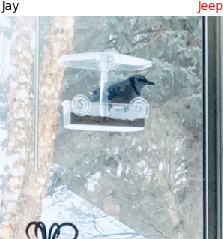}
\end{subfigure}
\caption{
Additional adversarially filtered examples from the \textsc{ImageNet-A} dataset. Examples are adversarially selected to cause classifier accuracy to degrade. The black text is the actual class, and the red text is a ResNet-50 prediction.
}\label{fig:imagenet-a}
\vspace{5pt}
\begin{subfigure}{.25\textwidth}
    \centering
    \includegraphics[width=0.98\textwidth]{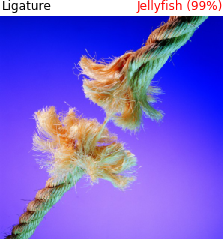}
\end{subfigure}%
\begin{subfigure}{.25\textwidth}
    \centering
    \includegraphics[width=0.98\textwidth]{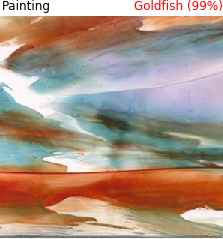}
\end{subfigure}%
\begin{subfigure}{.25\textwidth}
    \centering
    \includegraphics[width=0.98\textwidth]{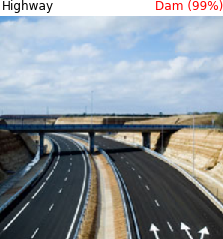}
\end{subfigure}%
\begin{subfigure}{.25\textwidth}
    \centering
    \includegraphics[width=0.98\textwidth]{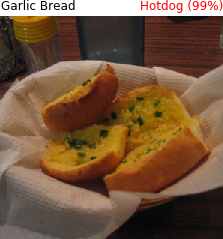}
\end{subfigure}
\caption{
Additional adversarially filtered examples from the \textsc{ImageNet-O} dataset. Examples are adversarially selected to cause out-of-distribution detection performance to degrade. Examples do not belong to ImageNet classes, and they are wrongly assigned highly confident predictions. The black text is the actual class, and the red text is a ResNet-50 prediction and the prediction confidence.
}\label{fig:imagenet-o}
\vspace{-10pt}
\end{figure*}

\section{\textsc{ImageNet-A} and \textsc{ImageNet-O}}\label{sec:design}
\subsection{Design}
\textsc{ImageNet-A} is a dataset of real-world adversarially filtered images that fool current ImageNet classifiers.
To find adversarially filtered examples, we first download numerous images related to an ImageNet class. Thereafter we delete the images that fixed ResNet-50 \cite{resnet} classifiers correctly predict. We chose ResNet-50 due to its widespread use. Later we show that examples which fool ResNet-50 reliably transfer to other unseen models. With the remaining incorrectly classified images, we manually select visually clear images.

Next, \textsc{ImageNet-O} is a dataset of adversarially filtered examples for ImageNet out-of-distribution detectors. To create this dataset, we download ImageNet-22K and delete examples from ImageNet-1K. With the remaining ImageNet-22K examples that do not belong to ImageNet-1K classes, we keep examples that are classified by a ResNet-50 as an ImageNet-1K class with high confidence. Then we manually select visually clear images.

Both datasets were manually constructed by graduate students over several months. This is because a large share of images contain multiple classes per image \cite{Stock2018ConvNetsAI}. Therefore, producing a dataset without multilabel images can be challenging with usual annotation techniques. To ensure images do not fall into more than one of the several hundred classes, we had graduate students memorize the classes in order to build a high-quality test set.

\noindent\textbf{\textsc{ImageNet-A} Class Restrictions.}\quad 
We select a $200$-class subset of ImageNet-1K's $1,000$ classes so that errors among these $200$ classes would be considered egregious \cite{imagenet}.
For instance, wrongly classifying Norwich terriers as Norfolk terriers does less to demonstrate faults in current classifiers than mistaking a Persian cat for a candle.
We additionally avoid rare classes such as ``snow leopard,'' classes that have changed much since 2012 such as ``iPod,'' coarse classes such as ``spiral,'' classes that are often image backdrops such as ``valley,'' and finally classes that tend to overlap such as ``honeycomb,'' ``bee,'' ``bee house,'' and ``bee eater''; ``eraser,'' ``pencil sharpener'' and ``pencil case''; ``sink,'' ``medicine cabinet,'' ``pill bottle'' and ``band-aid''; and so on. The $200$ \textsc{ImageNet-A} classes cover most broad categories spanned by ImageNet-1K; see the Supplementary Materials  
for the full class list.


\noindent\textbf{\textsc{ImageNet-A} Data Aggregation.}\quad The first step is to download many weakly labeled images. Fortunately, the website iNaturalist has millions of user-labeled images of animals, and Flickr has even more user-tagged images of objects. We download images related to each of the $200$ ImageNet classes by leveraging user-provided labels and tags. 
After exporting or scraping data from sites including iNaturalist, Flickr, and DuckDuckGo, we adversarially select images by removing examples that fail to fool our ResNet-50 models. Of the remaining images, we select low-confidence images and then ensure each image is valid through human review. If we only used the original ImageNet test set as a source rather than iNaturalist, Flickr, and DuckDuckGo, some classes would have zero images after the first round of filtration, as the original ImageNet test set is too small to contain hard adversarially filtered images.

We now describe this process in more detail. 
We use a small ensemble of ResNet-50s for filtering, one pre-trained on ImageNet-1K then fine-tuned on the $200$ class subset, and one pre-trained on ImageNet-1K where $200$ of its $1,000$ logits are used in classification. Both classifiers have similar accuracy on the $200$ clean test set classes from ImageNet-1K. The ResNet-50s perform 10-crop classification for each image, and should any crop be classified correctly by the ResNet-50s, the image is removed. If either ResNet-50 assigns greater than $15\%$ confidence to the correct class, the image is also removed; this is done so that adversarially filtered examples yield misclassifications with low confidence in the correct class, like in untargeted adversarial attacks. Now, some classification confusions are greatly over-represented, such as Persian cat and lynx. We would like \textsc{ImageNet-A} to have great variability in its types of errors and cause classifiers to have a dense confusion matrix. Consequently, we perform a second round of filtering to create a shortlist where each confusion only appears at most $15$ times. Finally, we manually select images from this shortlist in order to ensure \textsc{ImageNet-A} images are simultaneously valid, single-class, and high-quality. In all, the \textsc{ImageNet-A} dataset has $7,500$ adversarially filtered images.

As a specific example, we download $81,413$ dragonfly images from iNaturalist, and after running the ResNet-50 filter we have $8,925$ dragonfly images. In the algorithmically diversified shortlist, $1,452$ images remain. From this shortlist, $80$ dragonfly images are manually selected, but hundreds more could be selected if time allows.

The resulting images represent a substantial distribution shift, but images are still possible for humans to classify. The Fr\'echet Inception Distance (FID) \cite{Heusel2017GANsTB} enables us to determine whether \textsc{ImageNet-A} and ImageNet are not identically distributed. The FID between ImageNet’s validation and test set is approximately $0.99$, indicating that the distributions are highly similar. The FID between \textsc{ImageNet-A} and ImageNet’s validation set is $50.40$, and the FID between \textsc{ImageNet-A} and ImageNet's test set is approximately $50.25$, indicating that the distribution shift is large. Despite the shift, we estimate that our graduate students' \textsc{ImageNet-A} human accuracy rate is approximately $90\%$.

\noindent\textbf{\textsc{ImageNet-O} Class Restrictions.}\quad We again select a 200-class subset of ImageNet-1K's $1,000$ classes. These $200$ classes determine the in-distribution or the distribution that is considered usual. 
As before, the $200$ classes cover most broad categories spanned by ImageNet-1K; see the Supplementary Materials 
for the full class list. 

\begin{figure*}[t]
\vspace{-25pt}
\centering
\includegraphics[width=\textwidth]{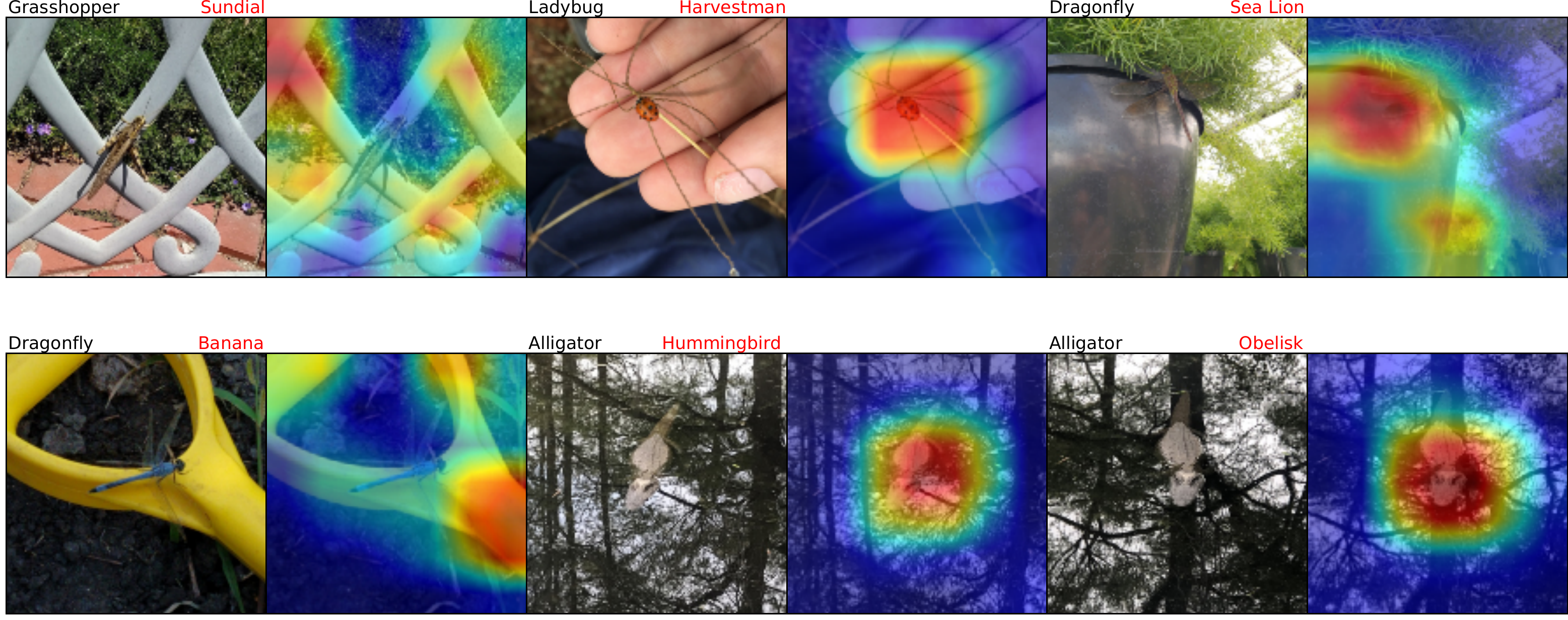}
\caption{
Examples from \textsc{ImageNet-A} demonstrating classifier failure modes. Adjacent to each natural image is its heatmap \cite{Selvaraju2019GradCAMVE}. Classifiers may use erroneous background cues for prediction. These failure modes are described in \Cref{sec:failures}.
}\label{fig:compilation}
\vspace{-10pt}
\end{figure*}

\noindent\textbf{\textsc{ImageNet-O} Data Aggregation.}\quad Our dataset for adversarial out-of-distribution detection is created by fooling ResNet-50 out-of-distribution detectors. The negative of the prediction confidence of a ResNet-50 ImageNet classifier serves as our anomaly score \cite{hendrycks17baseline}. Usually in-distribution examples produce higher confidence predictions than OOD examples, but we curate OOD examples that have high confidence predictions. To gather candidate adversarially filtered examples, we use the ImageNet-22K dataset with ImageNet-1K classes deleted. We choose the ImageNet-22K dataset since it was collected in the same way as ImageNet-1K. ImageNet-22K allows us to have coverage of numerous visual concepts and vary the distribution's semantics without unnatural or unwanted non-semantic data shift. After excluding ImageNet-1K images, we process the remaining ImageNet-22K images and keep the images which cause the ResNet-50 to have high confidence, or a low anomaly score. We then manually select a high-quality subset of the remaining images to create \textsc{ImageNet-O}. We suggest only training models with data from the $1,000$ ImageNet-1K classes, since the dataset becomes trivial if models train on ImageNet-22K. To our knowledge, this dataset is the first anomalous dataset curated for ImageNet models and enables researchers to study adversarial out-of-distribution detection. The \textsc{ImageNet-O} dataset has $2,000$ adversarially filtered examples since anomalies are rarer; this has the same number of examples per class as ImageNetV2 \cite{Recht2019DoIC}. While we use adversarial filtration to select images that are difficult for a fixed ResNet-50, we will show these examples straightforwardly transfer to unseen models. 

\subsection{Illustrative Failure Modes}\label{sec:failures}

Examples in \textsc{ImageNet-A} uncover numerous failure modes of modern convolutional neural networks. We describe our findings after having viewed tens of thousands of candidate adversarially filtered examples. Some of these failure modes may also explain poor \textsc{ImageNet-O} performance, but for simplicity we describe our observations with \textsc{ImageNet-A} examples.

Consider \Cref{fig:compilation}. The first two images suggest models may overgeneralize visual concepts. It may confuse metal with sundials, or thin radiating lines with harvestman bugs. We also observed that networks overgeneralize tricycles to bicycles and circles, digital clocks to keyboards and calculators, and more. We also observe that models may rely too heavily on color and texture, as shown with the dragonfly images. Since classifiers are taught to associate entire images with an object class, frequently appearing background elements may also become associated with a class, such as wood being associated with nails. Other examples include classifiers heavily associating hummingbird feeders with hummingbirds, leaf-covered tree branches being associated with the white-headed capuchin monkey class, snow being associated with shovels, and dumpsters with garbage trucks.
Additionally \Cref{fig:compilation} shows an American alligator swimming. With different frames, the classifier prediction varies erratically between classes that are semantically loose and separate. For other images of the swimming alligator, classifiers predict that the alligator is a cliff, lynx, and a fox squirrel. 
Assessing convolutional networks on \textsc{ImageNet-A} reveals that even state-of-the-art models have diverse and systematic failure modes.




\section{Experiments}\label{sec:experiments}
We show that adversarially filtered examples collected to fool fixed ResNet-50 models reliably transfer to other models, indicating that current convolutional neural networks have shared weaknesses and failure modes. In the following sections, we analyze whether robustness can be improved by using data augmentation, using more real labeled data, and using different architectures. For the first two sections, we analyze performance with a fixed architecture for comparability, and in the final section we observe performance with different architectures. 
First we define our metrics.

\begin{figure*}[t]
\centering
\vspace{-20pt}
\includegraphics[width=\textwidth]{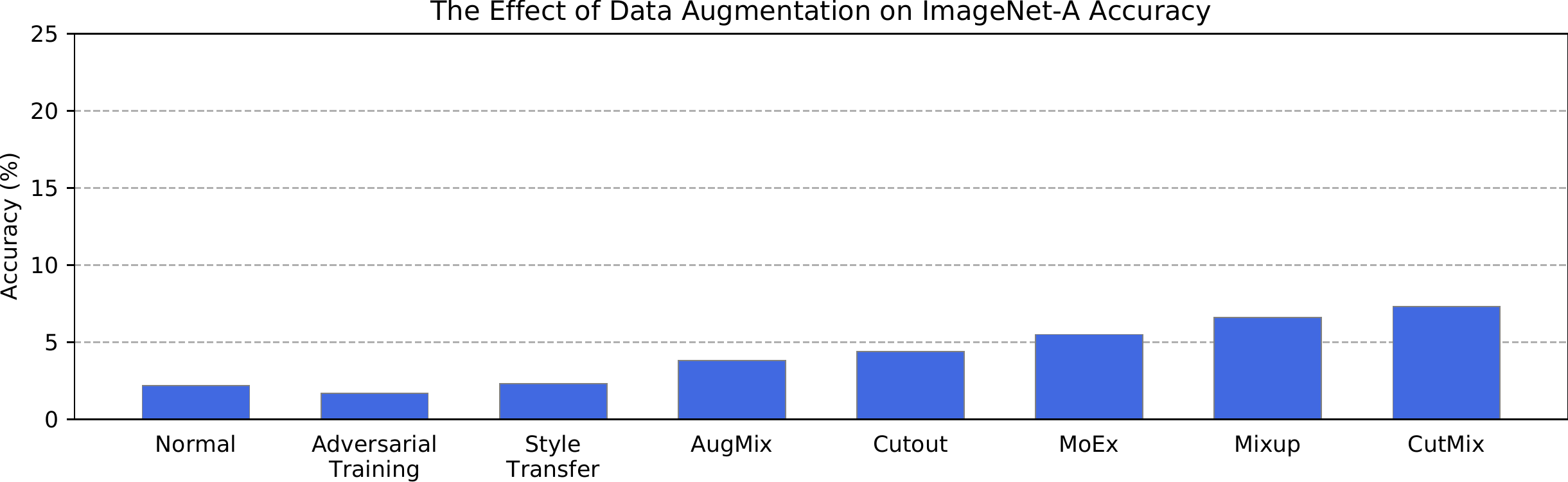}
\caption{Some data augmentation techniques hardly improve \textsc{ImageNet-A} accuracy. This demonstrates that \textsc{ImageNet-A} can expose previously unnoticed faults in proposed robustness methods which do well on synthetic distribution shifts \cite{hendrycks2019augmix}.}\label{fig:aug}
\vspace{-10pt}
\end{figure*}

\noindent\textbf{Metrics.} Our metric for assessing robustness to adversarially filtered examples for classifiers is the top-1 \emph{accuracy} on \textsc{ImageNet-A}. For reference, the top-1 accuracy on the 200 \textsc{ImageNet-A} classes using usual ImageNet images is usually greater than or equal to $90\%$ for ordinary classifiers.

Our metric for assessing out-of-distribution detection performance of \textsc{ImageNet-O} examples is the area under the precision-recall curve (\emph{AUPR}). This metric requires anomaly scores. Our anomaly score is the negative of the maximum softmax probabilities \cite{hendrycks17baseline} from a model that can classify the $200$ \textsc{ImageNet-O} classes. The maximum softmax probability detector is a long-standing baseline in OOD detection. We collect anomaly scores with the ImageNet validation examples for the said $200$ classes.
Then, we collect anomaly scores for the \textsc{ImageNet-O} examples. Higher performing OOD detectors would assign \textsc{ImageNet-O} examples lower confidences, or higher anomaly scores. With these anomaly scores, we can compute the area under the precision-recall curve \cite{auprbaseline}. Random chance levels for the AUPR is approximately $16.67\%$ with \textsc{ImageNet-O}, and the maximum AUPR is $100\%$.

\paragraph{Data Augmentation.}
We examine popular data augmentation techniques and note their effect on robustness. In this section we exclude \textsc{ImageNet-O} results, as the data augmentation techniques hardly help with out-of-distribution detection as well.
As a baseline, we train a new ResNet-50 from scratch and obtain $2.17\%$ accuracy on \textsc{ImageNet-A}.
Now, one purported way to increase robustness is through adversarial training, which makes models less sensitive to $\ell_p$ perturbations. We use the adversarially trained model from Wong et al., 2020 \cite{wong2020fast}, but accuracy decreases to $1.68\%$. Next, Geirhos et al., 2019 \cite{geirhos2019} propose making networks rely less on texture by training classifiers on images where textures are transferred from art pieces. They accomplish this by applying style transfer to ImageNet training images to create a stylized dataset, and models train on these images. While this technique is able to greatly increase robustness on synthetic corruptions \cite{hendrycks2019robustness}, Style Transfer increases \textsc{ImageNet-A} accuracy only $0.13\%$ over the ResNet-50 baseline. A recent data augmentation technique is AugMix \cite{hendrycks2019augmix}, which takes linear combinations of different data augmentations. This technique increases accuracy to $3.8\%$. Cutout augmentation \cite{Devries2017ImprovedRO} randomly occludes image regions and corresponds to $4.4\%$ accuracy. Moment Exchange (MoEx) \cite{Li2020OnFN} exchanges feature map moments between images, and this increases accuracy to $5.5\%$. Mixup \cite{Zhang2018mixupBE} trains networks on elementwise convex combinations of images and their interpolated labels; this technique increases accuracy to $6.6\%$. CutMix \cite{Yun2019CutMixRS} superimposes images regions within other images and yields $7.3\%$ accuracy. At best these data augmentations techniques improve accuracy by approximately $5\%$ over the baseline. Results are summarized in \Cref{fig:aug}. Although some data augmentation techniques are purported to greatly improve robustness to distribution shifts \cite{hendrycks2019augmix,Yin2019AFP}, their lackluster results on \textsc{ImageNet-A} show they do not improve robustness on some distribution shifts. Hence \textsc{ImageNet-A} can be used to verify whether techniques actually improve real-world robustness to distribution shift.

\begin{figure*}[t]
\vspace{-25pt}
\begin{subfigure}{.5\textwidth}
    \centering
    \includegraphics[width=0.99\textwidth]{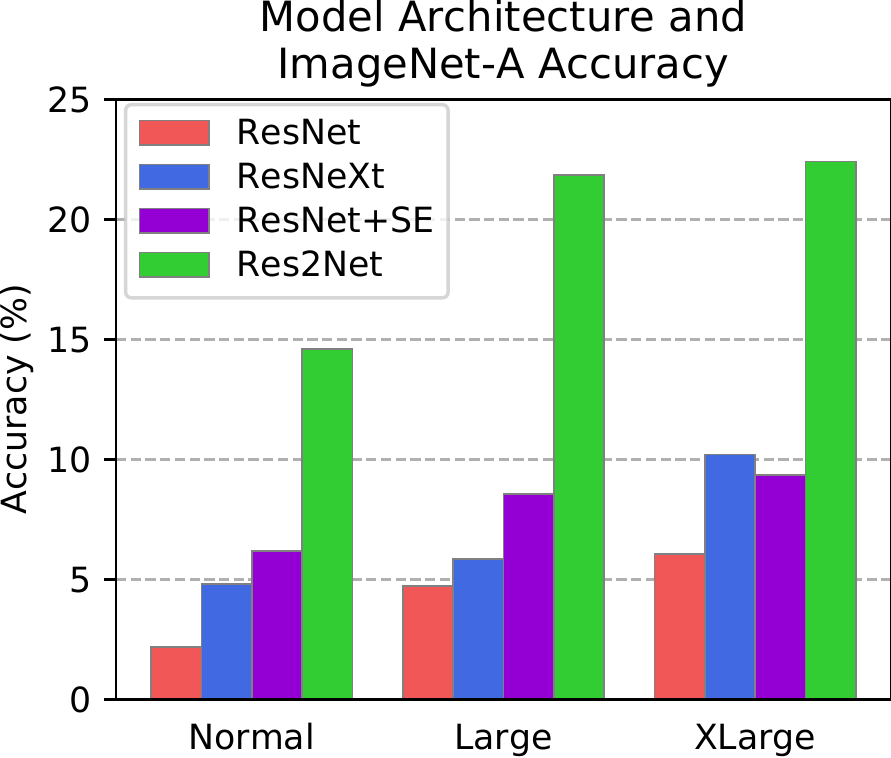}
\end{subfigure}%
\begin{subfigure}{.5\textwidth}
    \centering
    \includegraphics[width=0.99\textwidth]{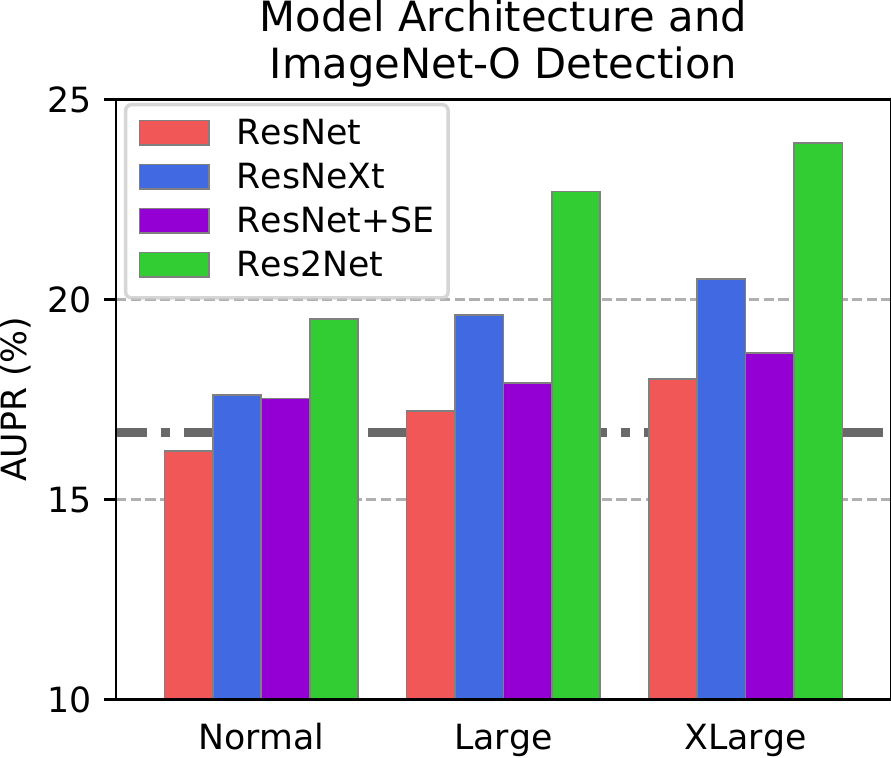}
\end{subfigure}%
\caption{Increasing model size and other architecture changes can greatly improve performance. Note Res2Net and ResNet+SE have a ResNet backbone. Normal model sizes are ResNet-50 and ResNeXt-50 ($32\times4$d), Large model sizes are ResNet-101 and ResNeXt-101 ($32\times4$d), and XLarge Model sizes are ResNet-152 and ($32\times8$d).}\label{fig:attn}
\vspace{-10pt}
\end{figure*}

\paragraph{More Labeled Data.}
One possible explanation for consistently low \textsc{ImageNet-A} accuracy is that all models are trained only with ImageNet-1K, and using additional data may resolve the problem. Bau et al., 2017 \cite{Bau2017NetworkDQ} argue that Places365 classifiers learn qualitatively distinct filters (e.g., they have more object detectors, fewer texture detectors in conv3) compared to ImageNet classifiers, so one may expect an error distribution less correlated with errors on ImageNet-A. To test this hypothesis we pre-train a ResNet-50 on Places365 \cite{zhou2017places}, a large-scale scene recognition dataset.
After fine-tuning the Places365 model on ImageNet-1K, we find that accuracy is $1.56\%$. Consequently, even though scene recognition models are purported to have qualitatively distinct features, this is not enough to improve \textsc{ImageNet-A} performance. Likewise, Places365 pre-training does not improve \textsc{ImageNet-O} detection, as its AUPR is $14.88\%$. Next, we see whether labeled data from \textsc{ImageNet-A} itself can help. We take baseline ResNet-50 with $2.17\%$ \textsc{ImageNet-A} accuracy and fine-tune it on $80\%$ of \textsc{ImageNet-A}. This leads to no clear improvement on the remaining $20\%$ of \textsc{ImageNet-A} since the top-1 and top-5 accuracies are below $2\%$ and $5\%$, respectively.

Last, we pre-train using an order of magnitude more training data with ImageNet-21K. This dataset contains approximately $21,000$ classes and approximately $14$ million images. To our knowledge this is the largest publicly available database of labeled natural images. Using a ResNet-50 pretrained on ImageNet-21K, we fine-tune the model on ImageNet-1K and attain $11.41\%$ accuracy on \textsc{ImageNet-A}, a $9.24\%$ increase. Likewise, the AUPR for \textsc{ImageNet-O} improves from $16.20\%$ to $21.86\%$, although this improvement is less significant since \textsc{ImageNet-O} images overlap with ImageNet-21K images. Academic researchers rarely use datasets larger than ImageNet due to computational costs, using more data has limitations. An order of magnitude increase in labeled training data can provide some improvements in accuracy, though we now show that architecture changes provide greater improvements.

\paragraph{Architectural Changes.}
We find that model architecture can play a large role in \textsc{ImageNet-A} accuracy and \textsc{ImageNet-O} detection performance.
Simply increasing the width and number of layers of a network is sufficient to automatically impart more \textsc{ImageNet-A} accuracy and \textsc{ImageNet-O} OOD detection performance. Increasing network capacity has been shown to improve performance on $\ell_p$ adversarial examples \cite{kurakin}, common corruptions \cite{hendrycks2019robustness}, and now also improves performance for adversarially filtered images. For example, a ResNet-50's top-1 accuracy and AUPR is $2.17\%$ and $16.2\%$, respectively, while a ResNet-152 obtains $6.1\%$ top-1 accuracy and $18.0\%$ AUPR. Another architecture change that reliably helps is using the grouped convolutions found in ResNeXts \cite{resnext}. A ResNeXt-50 ($32\times4$d) obtains a $4.81\%$ top1 \textsc{ImageNet-A} accuracy and a $17.60\%$ \textsc{ImageNet-O} AUPR.

Another useful architecture change is self-attention.
Convolutional neural networks with self-attention \cite{Hu2018GatherExciteE} are designed to better capture long-range dependencies and interactions across an image. We consider the self-attention technique called Squeeze-and-Excitation (SE) \cite{Hu2018SqueezeandExcitationN}, which won the final ImageNet competition in 2017. A ResNet-50 with Squeeze-and-Excitation attains $6.17\%$ accuracy. However, for larger ResNets, self-attention does little to improve \textsc{ImageNet-O} detection.

We consider the ResNet-50 architecture with its residual blocks exchanged with recently introduced Res2Net v1b blocks \cite{Gao2019Res2NetAN}. This change increases accuracy to $14.59\%$ and the AUPR to $19.5\%$. A ResNet-152 with Res2Net v1b blocks attains $22.4\%$ accuracy and $23.9\%$ AUPR. Compared to data augmentation or an order of magnitude more labeled training data, some architectural changes can provide far more robustness gains. Consequently future improvements to model architectures is a promising path towards greater robustness.

We now assess performance on a completely different architecture which does not use convolutions, vision Transformers \cite{Dosovitskiy2020AnII}.
We evaluate with DeiT \cite{touvron2020deit}, a vision Transformer trained on ImageNet-1K with aggressive data augmentation such as Mixup. \emph{Even for vision Transformers, we find that ImageNet-A and ImageNet-O examples successfully transfer.} In particular, a DeiT-small vision Transformer gets 19.0\% on \textsc{ImageNet-A} and has a similar number of parameters to a Res2Net-50, which has 14.6\% accuracy. This might be explained by DeiT's use of Mixup, however, which provided a 4\% ImageNet-A accuracy boost for ResNets. The \textsc{ImageNet-O} AUPR for the Transformer is 20.9\%, while the Res2Net gets 19.5\%. Larger DeiT models do better, as a DeiT-base gets 28.2\% accuracy on \textsc{ImageNet-A} and 24.8\% AUPR on \textsc{ImageNet}. Consequently, our datasets transfer to vision Transformers and performance for both tasks remains far from the ceiling.

\section{Conclusion}

We found it is possible to improve performance on our datasets with data augmentation, pretraining data, and architectural changes. We found that our examples transferred to all tested models, including vision Transformers which do not use convolution operations. Results indicate that improving performance on \textsc{ImageNet-A} and \textsc{ImageNet-O} is possible but difficult. Our challenging ImageNet test sets serve as measures of performance under distribution shift---an important research aim as models are deployed in increasingly precarious real-world environments.

\newpage
{\small
\bibliographystyle{styles/cvpr/ieee_fullname}
\bibliography{main.bib}
}
\newpage
\section{Appendix}
\begin{table*}[t]
\vspace{-10pt}
\begin{center}
\begin{tabular}{lcc}
\hline
                        & ImageNet-A (Acc \%) & ImageNet-O (AUPR \%) \\
                        \hline
AlexNet                 & 1.77  & 15.44     \\
SqueezeNet1.1           & 1.12  & 15.31     \\
VGG16                   & 2.63  & 16.58     \\
VGG19                   & 2.11  & 16.80     \\
VGG19+BN                & 2.95  & 16.57     \\
DenseNet121             & 2.16  & 16.11     \\
\hline
ResNet-18               & 1.15  & 15.23     \\
ResNet-34               & 1.87  & 16.00     \\
ResNet-50               & 2.17  & 16.20     \\
ResNet-101              & 4.72  & 17.20     \\
ResNet-152              & 6.05  & 18.00     \\
\hline
ResNet-50+Squeeze-and-Excite            & 6.17  & 17.52     \\
ResNet-101+Squeeze-and-Excite           & 8.55  & 17.91     \\
ResNet-152+Squeeze-and-Excite           & 9.35  & 18.65     \\
\hline
ResNet-50+DeVries Confidence Branch & 0.35 & 14.34 \\
ResNet-50+Rotation Prediction Branch & 2.17 & 16.20 \\
\hline
Res2Net-50 (v1b)        & 14.59 & 19.50     \\
Res2Net-101 (v1b)       & 21.84 & 22.69     \\
Res2Net-152 (v1b)       & 22.4  & 23.90     \\
\hline
ResNeXt-50 ($32\times4$d)   & 4.81  & 17.60 \\
ResNeXt-101 ($32\times4$d)  & 5.85  & 19.60 \\
ResNeXt-101 ($32\times8$d)   & 10.2  & 20.51 \\
\hline
DPN 68                  & 3.53  & 17.78     \\
DPN 98                  & 9.15  & 21.10     \\
\hline
DeiT-tiny        & 7.25 & 17.4 \\
DeiT-small        & 19.1 & 20.9 \\
DeiT-base        & 28.2 & 24.8 \\
\hline
\end{tabular}
\end{center}
\caption{Expanded \textsc{ImageNet-A} and \textsc{ImageNet-O} architecture results. Note \textsc{ImageNet-O} performance is improving more slowly.}\label{tab:fullarch}
\vspace{-10pt}
\end{table*}

\section{Expanded Results}

\subsection{Full Architecture Results}
Full results with various architectures are in \Cref{tab:fullarch}.

\subsection{More OOD Detection Results and Background}

Works in out-of-distribution detection frequently use the maximum softmax baseline to detect out-of-distribution examples \cite{hendrycks17baseline}. Before neural networks, using the reject option or a $k+1$st class was somewhat common \cite{Bartlett2008ClassificationWA}, but with neural networks it requires auxiliary anomalous training data. New neural methods that utilize auxiliary anomalous training data, such as Outlier Exposure \cite{hendrycks2019oe}, do not use the reject option and still utilize the maximum softmax probability. We do not use Outlier Exposure since that paper's authors were unable to get their technique to work on ImageNet-1K with $224\times224$ images, though they were able to get it work on Tiny ImageNet which has $64\times64$ images. We do not use ODIN since it requires tuning hyperparameters directly using out-of-distribution data, a criticized practice \cite{hendrycks2019oe}.


We evaluate three additional out-of-distribution detection methods, though none substantially improve performance. We evaluate method of \cite{Devries2018LearningCF}, which trains an auxiliary branch to represent the model confidence. Using a ResNet trained from scratch, we find this gets a $14.3\%$ AUPR, around 2\% less than the MSP baseline. Next we use the recent Maximum Logit detector \cite{Hendrycks2020ScalingOD}. With 
DenseNet-121 the AUPR decreases from $16.1\%$ (MSP) to $15.8\%$ (Max Logit), while with ResNeXt-101 ($32\times8$d) the AUPR of $20.5\%$ increases to $20.6\%$. Across over 10 models we found the MaxLogit technique to be slightly worse. Finally, we evaluate the utility of self-supervised auxiliary objectives for OOD detection. The rotation prediction anomaly detector \cite{Hendrycks2019UsingSL} was shown to help improve detection performance for near-distribution yet still out-of-class examples, and with this auxiliary objective the AUPR for ResNet-50 does not change; it is $16.2\%$ with the rotation prediction and $16.2\%$ with the MSP. Note this method requires training the network and does not work out-of-the-box. 

\begin{figure}[]
\centering
	\includegraphics[width=0.16\textwidth]{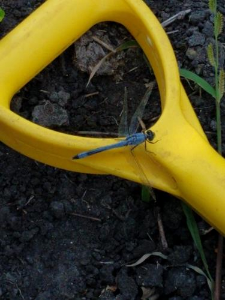}%
	\includegraphics[width=0.16\textwidth]{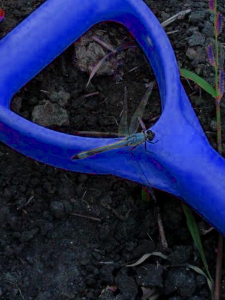}%
	\includegraphics[width=0.16\textwidth]{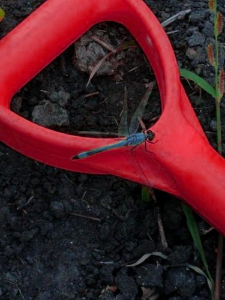}
	\caption{A demonstration of color sensitivity. While the leftmost image is classified as ``banana'' with high confidence, the images with modified color are correctly classified. Not only would we like models to be more accurate, we would like them to be calibrated if they wrong.}\label{fig:dragonfly}
\end{figure}

\subsection{Calibration}\label{app:calibration}
In this section we show \textsc{ImageNet-A} calibration results.


\noindent\textbf{Uncertainty Metrics.} \quad The \textit{$\ell_2$ Calibration Error} is how we measure miscalibration. We would like classifiers that can reliably forecast their accuracy. Concretely, we want classifiers which give examples 60\% confidence to be correct 60\% of the time. We judge a classifier's miscalibration with the $\ell_2$ Calibration Error \cite{kumar2019calibration}.


\begin{figure}[t]
	\centering
	\includegraphics[width=0.48\textwidth]{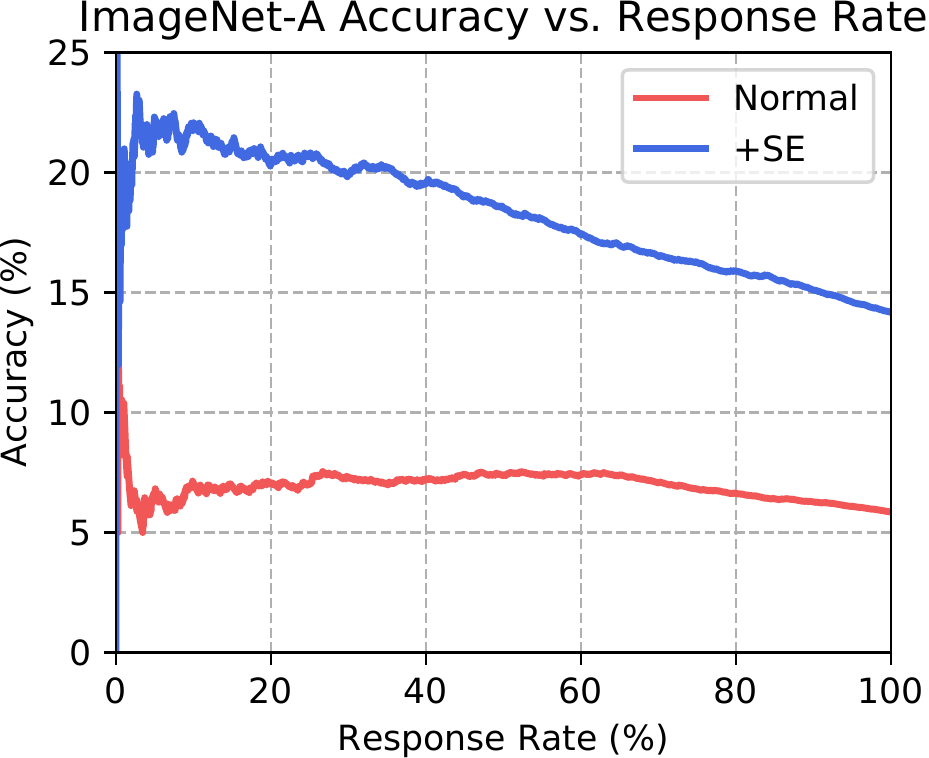}
	\caption{The Response Rate Accuracy curve for a ResNeXt-101 (32$\times$4d) with and without Squeeze-and-Excitation (SE). The Response Rate is the percent classified. The accuracy at a $n$\% response rate is the accuracy on the $n$\% of examples where the classifier is most confident.
	}
	\label{fig:rra}
\end{figure}

Our second uncertainty estimation metric is the \textit{Area Under the Response Rate Accuracy Curve (AURRA).} Responding only when confident is often preferable to predicting falsely.
In these experiments, we allow classifiers to respond to a subset of the test set and abstain from predicting the rest. Classifiers with quality uncertainty estimates should be capable identifying examples it is likely to predict falsely and abstain. If a classifier is required to abstain from predicting on 90\% of the test set, or equivalently respond to the remaining 10\% of the test set, then we should like the classifier's uncertainty estimates to separate correctly and falsely classified examples and have high accuracy on the selected 10\%. At a fixed response rate, we should like the accuracy to be as high as possible. At a 100\% response rate, the classifier accuracy is the usual test set accuracy. We vary the response rates and compute the corresponding accuracies to obtain the Response Rate Accuracy (RRA) curve. The area under the Response Rate Accuracy curve is the AURRA. To compute the AURRA in this paper, we use the maximum softmax probability. For response rate $p$, we take the $p$ fraction of examples with highest maximum softmax probability. If the response rate is 10\%, we select the top 10\% of examples with the highest confidence and compute the accuracy on these examples. An example RRA curve is in \Cref{fig:rra} .\looseness=-1


\begin{figure}[]
\begin{subfigure}{.48\textwidth}
    \centering
    \includegraphics[width=\textwidth]{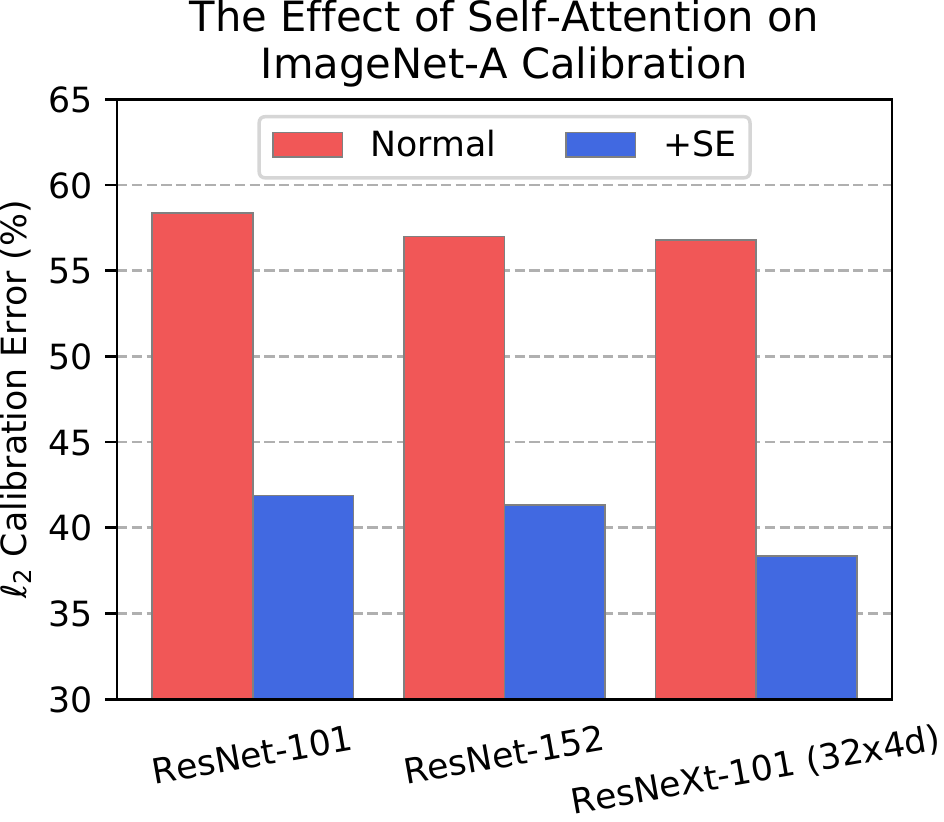}
\end{subfigure}
\begin{subfigure}{.48\textwidth}
    \centering
    \includegraphics[width=\textwidth]{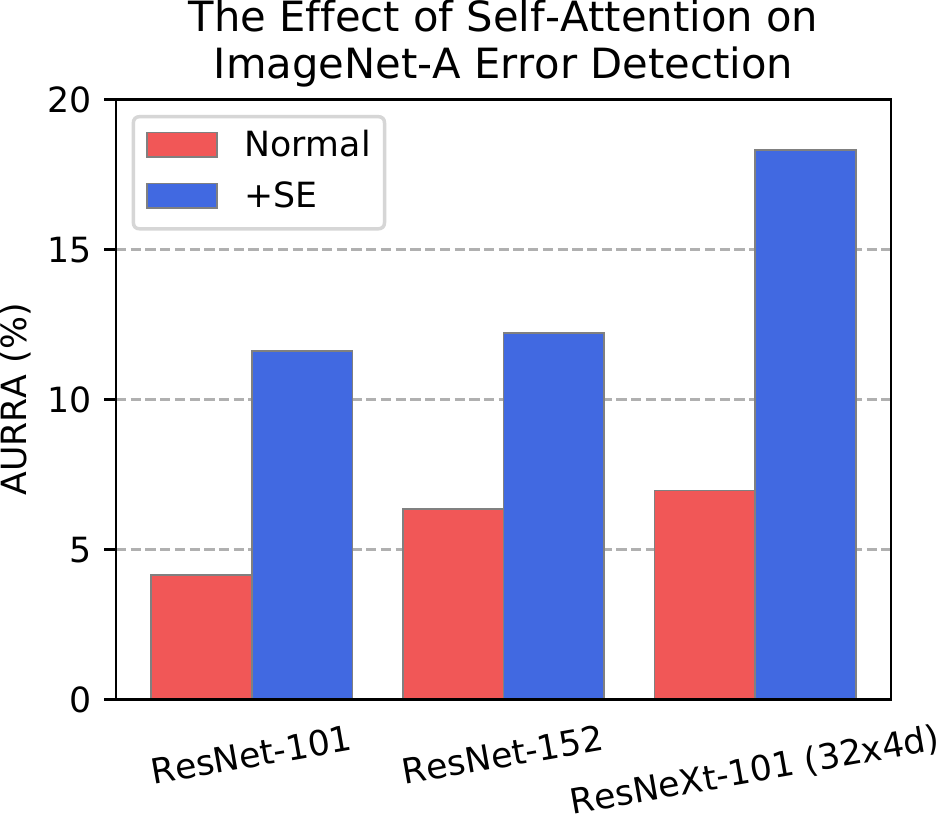}
\end{subfigure}%
\caption{Self-attention's influence on \textsc{ImageNet-A} $\ell_2$ calibration and error detection.}
\end{figure}


\begin{figure}
\begin{subfigure}{.48\textwidth}
    \centering
    \includegraphics[width=\textwidth]{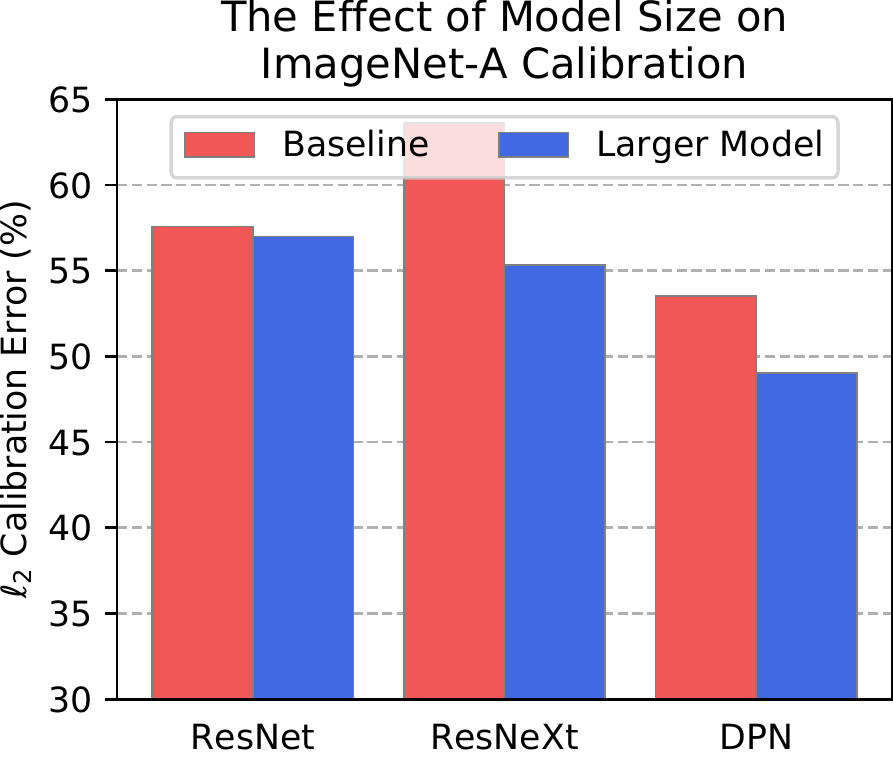}
\end{subfigure}
\begin{subfigure}{.48\textwidth}
    \centering
    \includegraphics[width=\textwidth]{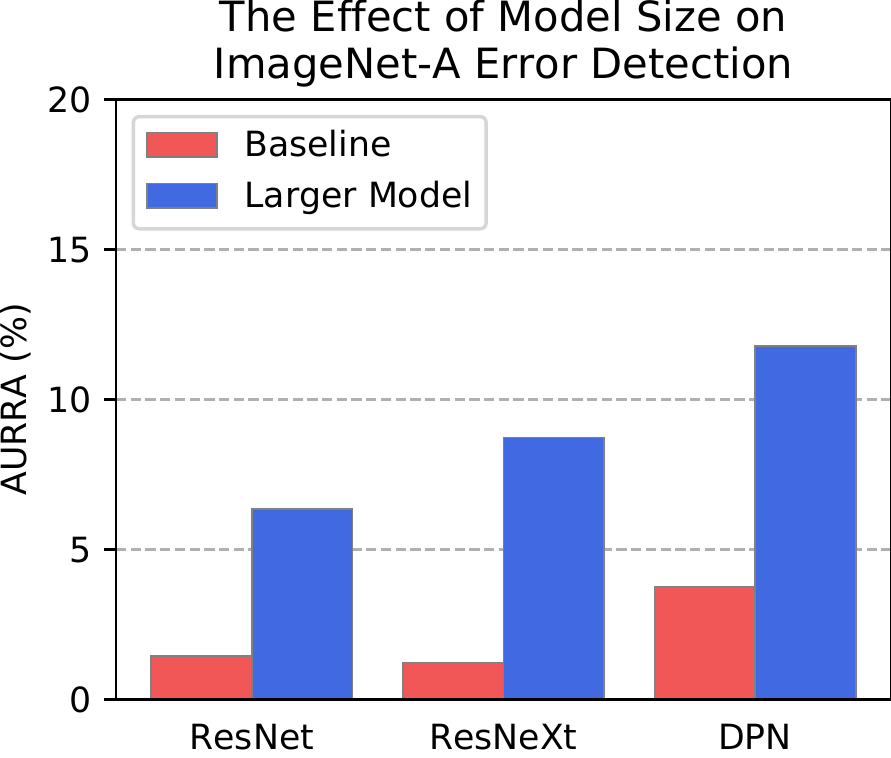}
\end{subfigure}%
\caption{Model size's influence on \textsc{ImageNet-A} $\ell_2$ calibration and error detection.}
\end{figure}

\newpage
\newpage

\section{\textsc{ImageNet-A} Classes}\label{app:classes}
The 200 ImageNet classes that we selected for \textsc{ImageNet-A} are as follows.
goldfish, \quad 
great white shark, \quad 
hammerhead, \quad 
stingray, \quad 
hen, \quad 
ostrich, \quad 
goldfinch, \quad 
junco, \quad 
bald eagle, \quad 
vulture, \quad 
newt, \quad 
axolotl, \quad 
tree frog, \quad 
iguana, \quad 
African chameleon, \quad 
cobra, \quad 
scorpion, \quad 
tarantula, \quad 
centipede, \quad 
peacock, \quad 
lorikeet, \quad 
hummingbird, \quad 
toucan, \quad 
duck, \quad 
goose, \quad 
black swan, \quad 
koala, \quad 
jellyfish, \quad 
snail, \quad 
lobster, \quad 
hermit crab, \quad 
flamingo, \quad 
american egret, \quad 
pelican, \quad 
king penguin, \quad 
grey whale, \quad 
killer whale, \quad 
sea lion, \quad 
chihuahua, \quad 
shih tzu, \quad 
afghan hound, \quad 
basset hound, \quad 
beagle, \quad 
bloodhound, \quad 
italian greyhound, \quad 
whippet, \quad 
weimaraner, \quad 
yorkshire terrier, \quad 
boston terrier, \quad 
scottish terrier, \quad 
west highland white terrier, \quad 
golden retriever, \quad 
labrador retriever, \quad 
cocker spaniels, \quad 
collie, \quad 
border collie, \quad 
rottweiler, \quad 
german shepherd dog, \quad 
boxer, \quad 
french bulldog, \quad 
saint bernard, \quad 
husky, \quad 
dalmatian, \quad 
pug, \quad 
pomeranian, \quad 
chow chow, \quad 
pembroke welsh corgi, \quad 
toy poodle, \quad 
standard poodle, \quad 
timber wolf, \quad 
hyena, \quad 
red fox, \quad 
tabby cat, \quad 
leopard, \quad 
snow leopard, \quad 
lion, \quad 
tiger, \quad 
cheetah, \quad 
polar bear, \quad 
meerkat, \quad 
ladybug, \quad 
fly, \quad 
bee, \quad 
ant, \quad 
grasshopper, \quad 
cockroach, \quad 
mantis, \quad 
dragonfly, \quad 
monarch butterfly, \quad 
starfish, \quad 
wood rabbit, \quad 
porcupine, \quad 
fox squirrel, \quad 
beaver, \quad 
guinea pig, \quad 
zebra, \quad 
pig, \quad 
hippopotamus, \quad 
bison, \quad 
gazelle, \quad 
llama, \quad 
skunk, \quad 
badger, \quad 
orangutan, \quad 
gorilla, \quad 
chimpanzee, \quad 
gibbon, \quad 
baboon, \quad 
panda, \quad 
eel, \quad 
clown fish, \quad 
puffer fish, \quad 
accordion, \quad 
ambulance, \quad 
assault rifle, \quad 
backpack, \quad 
barn, \quad 
wheelbarrow, \quad 
basketball, \quad 
bathtub, \quad 
lighthouse, \quad 
beer glass, \quad 
binoculars, \quad 
birdhouse, \quad 
bow tie, \quad 
broom, \quad 
bucket, \quad 
cauldron, \quad 
candle, \quad 
cannon, \quad 
canoe, \quad 
carousel, \quad 
castle, \quad 
mobile phone, \quad 
cowboy hat, \quad 
electric guitar, \quad 
fire engine, \quad 
flute, \quad 
gasmask, \quad 
grand piano, \quad 
guillotine, \quad 
hammer, \quad 
harmonica, \quad 
harp, \quad 
hatchet, \quad 
jeep, \quad 
joystick, \quad 
lab coat, \quad 
lawn mower, \quad 
lipstick, \quad 
mailbox, \quad 
missile, \quad 
mitten, \quad 
parachute, \quad 
pickup truck, \quad 
pirate ship, \quad 
revolver, \quad 
rugby ball, \quad 
sandal, \quad 
saxophone, \quad 
school bus, \quad 
schooner, \quad 
shield, \quad 
soccer ball, \quad 
space shuttle, \quad 
spider web, \quad 
steam locomotive, \quad 
scarf, \quad 
submarine, \quad 
tank, \quad 
tennis ball, \quad 
tractor, \quad 
trombone, \quad 
vase, \quad 
violin, \quad 
military aircraft, \quad 
wine bottle, \quad 
ice cream, \quad 
bagel, \quad 
pretzel, \quad 
cheeseburger, \quad 
hotdog, \quad 
cabbage, \quad 
broccoli, \quad 
cucumber, \quad 
bell pepper, \quad 
mushroom, \quad 
Granny Smith, \quad 
strawberry, \quad 
lemon, \quad 
pineapple, \quad 
banana, \quad 
pomegranate, \quad 
pizza, \quad 
burrito, \quad 
espresso, \quad 
volcano, \quad 
baseball player, \quad 
scuba diver, \quad 
acorn, \quad 

n01443537, \quad 
n01484850, \quad 
n01494475, \quad 
n01498041, \quad 
n01514859, \quad 
n01518878, \quad 
n01531178, \quad 
n01534433, \quad 
n01614925, \quad 
n01616318, \quad 
n01630670, \quad 
n01632777, \quad 
n01644373, \quad 
n01677366, \quad 
n01694178, \quad 
n01748264, \quad 
n01770393, \quad 
n01774750, \quad 
n01784675, \quad 
n01806143, \quad 
n01820546, \quad 
n01833805, \quad 
n01843383, \quad 
n01847000, \quad 
n01855672, \quad 
n01860187, \quad 
n01882714, \quad 
n01910747, \quad 
n01944390, \quad 
n01983481, \quad 
n01986214, \quad 
n02007558, \quad 
n02009912, \quad 
n02051845, \quad 
n02056570, \quad 
n02066245, \quad 
n02071294, \quad 
n02077923, \quad 
n02085620, \quad 
n02086240, \quad 
n02088094, \quad 
n02088238, \quad 
n02088364, \quad 
n02088466, \quad 
n02091032, \quad 
n02091134, \quad 
n02092339, \quad 
n02094433, \quad 
n02096585, \quad 
n02097298, \quad 
n02098286, \quad 
n02099601, \quad 
n02099712, \quad 
n02102318, \quad 
n02106030, \quad 
n02106166, \quad 
n02106550, \quad 
n02106662, \quad 
n02108089, \quad 
n02108915, \quad 
n02109525, \quad 
n02110185, \quad 
n02110341, \quad 
n02110958, \quad 
n02112018, \quad 
n02112137, \quad 
n02113023, \quad 
n02113624, \quad 
n02113799, \quad 
n02114367, \quad 
n02117135, \quad 
n02119022, \quad 
n02123045, \quad 
n02128385, \quad 
n02128757, \quad 
n02129165, \quad 
n02129604, \quad 
n02130308, \quad 
n02134084, \quad 
n02138441, \quad 
n02165456, \quad 
n02190166, \quad 
n02206856, \quad 
n02219486, \quad 
n02226429, \quad 
n02233338, \quad 
n02236044, \quad 
n02268443, \quad 
n02279972, \quad 
n02317335, \quad 
n02325366, \quad 
n02346627, \quad 
n02356798, \quad 
n02363005, \quad 
n02364673, \quad 
n02391049, \quad 
n02395406, \quad 
n02398521, \quad 
n02410509, \quad 
n02423022, \quad 
n02437616, \quad 
n02445715, \quad 
n02447366, \quad 
n02480495, \quad 
n02480855, \quad 
n02481823, \quad 
n02483362, \quad 
n02486410, \quad 
n02510455, \quad 
n02526121, \quad 
n02607072, \quad 
n02655020, \quad 
n02672831, \quad 
n02701002, \quad 
n02749479, \quad 
n02769748, \quad 
n02793495, \quad 
n02797295, \quad 
n02802426, \quad 
n02808440, \quad 
n02814860, \quad 
n02823750, \quad 
n02841315, \quad 
n02843684, \quad 
n02883205, \quad 
n02906734, \quad 
n02909870, \quad 
n02939185, \quad 
n02948072, \quad 
n02950826, \quad 
n02951358, \quad 
n02966193, \quad 
n02980441, \quad 
n02992529, \quad 
n03124170, \quad 
n03272010, \quad 
n03345487, \quad 
n03372029, \quad 
n03424325, \quad 
n03452741, \quad 
n03467068, \quad 
n03481172, \quad 
n03494278, \quad 
n03495258, \quad 
n03498962, \quad 
n03594945, \quad 
n03602883, \quad 
n03630383, \quad 
n03649909, \quad 
n03676483, \quad 
n03710193, \quad 
n03773504, \quad 
n03775071, \quad 
n03888257, \quad 
n03930630, \quad 
n03947888, \quad 
n04086273, \quad 
n04118538, \quad 
n04133789, \quad 
n04141076, \quad 
n04146614, \quad 
n04147183, \quad 
n04192698, \quad 
n04254680, \quad 
n04266014, \quad 
n04275548, \quad 
n04310018, \quad 
n04325704, \quad 
n04347754, \quad 
n04389033, \quad 
n04409515, \quad 
n04465501, \quad 
n04487394, \quad 
n04522168, \quad 
n04536866, \quad 
n04552348, \quad 
n04591713, \quad 
n07614500, \quad 
n07693725, \quad 
n07695742, \quad 
n07697313, \quad 
n07697537, \quad 
n07714571, \quad 
n07714990, \quad 
n07718472, \quad 
n07720875, \quad 
n07734744, \quad 
n07742313, \quad 
n07745940, \quad 
n07749582, \quad 
n07753275, \quad 
n07753592, \quad 
n07768694, \quad 
n07873807, \quad 
n07880968, \quad 
n07920052, \quad 
n09472597, \quad 
n09835506, \quad 
n10565667, \quad 
n12267677, \quad 

\noindent`Stingray;' `goldfinch, Carduelis carduelis;' `junco, snowbird;' `robin, American robin, Turdus migratorius;' `jay;' `bald eagle, American eagle, Haliaeetus leucocephalus;' `vulture;' `eft;' `bullfrog, Rana catesbeiana;' `box turtle, box tortoise;' `common iguana, iguana, Iguana iguana;' `agama;' `African chameleon, Chamaeleo chamaeleon;' `American alligator, Alligator mississipiensis;' `garter snake, grass snake;' `harvestman, daddy longlegs, Phalangium opilio;' `scorpion;' `tarantula;' `centipede;' `sulphur-crested cockatoo, Kakatoe galerita, Cacatua galerita;' `lorikeet;' `hummingbird;' `toucan;' `drake;' `goose;' `koala, koala bear, kangaroo bear, native bear, Phascolarctos cinereus;' `jellyfish;' `sea anemone, anemone;' `flatworm, platyhelminth;' `snail;' `crayfish, crawfish, crawdad, crawdaddy;' `hermit crab;' `flamingo;' `American egret, great white heron, Egretta albus;' `oystercatcher, oyster catcher;' `pelican;' `sea lion;' `Chihuahua;' `golden retriever;' `Rottweiler;' `German shepherd, German shepherd dog, German police dog, alsatian;' `pug, pug-dog;' `red fox, Vulpes vulpes;' `Persian cat;' `lynx, catamount;' `lion, king of beasts, Panthera leo;' `American black bear, black bear, Ursus americanus, Euarctos americanus;' `mongoose;' `ladybug, ladybeetle, lady beetle, ladybird, ladybird beetle;' `rhinoceros beetle;' `weevil;' `fly;' `bee;' `ant, emmet, pismire;' `grasshopper, hopper;' `walking stick, walkingstick, stick insect;' `cockroach, roach;' `mantis, mantid;' `leafhopper;' `dragonfly, darning needle, devil's darning needle, sewing needle, snake feeder, snake doctor, mosquito hawk, skeeter hawk;' `monarch, monarch butterfly, milkweed butterfly, Danaus plexippus;' `cabbage butterfly;' `lycaenid, lycaenid butterfly;' `starfish, sea star;' `wood rabbit, cottontail, cottontail rabbit;' `porcupine, hedgehog;' `fox squirrel, eastern fox squirrel, Sciurus niger;' `marmot;' `bison;' `skunk, polecat, wood pussy;' `armadillo;' `baboon;' `capuchin, ringtail, Cebus capucinus;' `African elephant, Loxodonta africana;' `puffer, pufferfish, blowfish, globefish;' `academic gown, academic robe, judge's robe;' `accordion, piano accordion, squeeze box;' `acoustic guitar;' `airliner;' `ambulance;' `apron;' `balance beam, beam;' `balloon;' `banjo;' `barn;' `barrow, garden cart, lawn cart, wheelbarrow;' `basketball;' `beacon, lighthouse, beacon light, pharos;' `beaker;' `bikini, two-piece;' `bow;' `bow tie, bow-tie, bowtie;' `breastplate, aegis, egis;' `broom;' `candle, taper, wax light;' `canoe;' `castle;' `cello, violoncello;' `chain;' `chest;' `Christmas stocking;' `cowboy boot;' `cradle;' `dial telephone, dial phone;' `digital clock;' `doormat, welcome mat;' `drumstick;' `dumbbell;' `envelope;' `feather boa, boa;' `flagpole, flagstaff;' `forklift;' `fountain;' `garbage truck, dustcart;' `goblet;' `go-kart;' `golfcart, golf cart;' `grand piano, grand;' `hand blower, blow dryer, blow drier, hair dryer, hair drier;' `iron, smoothing iron;' `jack-o'-lantern;' `jeep, landrover;' `kimono;' `lighter, light, igniter, ignitor;' `limousine, limo;' `manhole cover;' `maraca;' `marimba, xylophone;' `mask;' `mitten;' `mosque;' `nail;' `obelisk;' `ocarina, sweet potato;' `organ, pipe organ;' `parachute, chute;' `parking meter;' `piggy bank, penny bank;' `pool table, billiard table, snooker table;' `puck, hockey puck;' `quill, quill pen;' `racket, racquet;' `reel;' `revolver, six-gun, six-shooter;' `rocking chair, rocker;' `rugby ball;' `saltshaker, salt shaker;' `sandal;' `sax, saxophone;' `school bus;' `schooner;' `sewing machine;' `shovel;' `sleeping bag;' `snowmobile;' `snowplow, snowplough;' `soap dispenser;' `spatula;' `spider web, spider's web;' `steam locomotive;' `stethoscope;' `studio couch, day bed;' `submarine, pigboat, sub, U-boat;' `sundial;' `suspension bridge;' `syringe;' `tank, army tank, armored combat vehicle, armoured combat vehicle;' `teddy, teddy bear;' `toaster;' `torch;' `tricycle, trike, velocipede;' `umbrella;' `unicycle, monocycle;' `viaduct;' `volleyball;' `washer, automatic washer, washing machine;' `water tower;' `wine bottle;' `wreck;' `guacamole;' `pretzel;' `cheeseburger;' `hotdog, hot dog, red hot;' `broccoli;' `cucumber, cuke;' `bell pepper;' `mushroom;' `lemon;' `banana;' `custard apple;' `pomegranate;' `carbonara;' `bubble;' `cliff, drop, drop-off;' `volcano;' `ballplayer, baseball player;' `rapeseed;' `yellow lady's slipper, yellow lady-slipper, Cypripedium calceolus, Cypripedium parviflorum;' `corn;' `acorn.'

Their WordNet IDs are as follows.

\noindent n01498041, \quad n01531178, \quad n01534433, \quad n01558993, \quad n01580077, \quad n01614925, \quad n01616318, \quad n01631663, \quad n01641577, \quad n01669191, \quad n01677366, \quad n01687978, \quad n01694178, \quad n01698640, \quad n01735189, \quad n01770081, \quad n01770393, \quad n01774750, \quad n01784675, \quad n01819313, \quad n01820546, \quad n01833805, \quad n01843383, \quad n01847000, \quad n01855672, \quad n01882714, \quad n01910747, \quad n01914609, \quad n01924916, \quad n01944390, \quad n01985128, \quad n01986214, \quad n02007558, \quad n02009912, \quad n02037110, \quad n02051845, \quad n02077923, \quad n02085620, \quad n02099601, \quad n02106550, \quad n02106662, \quad n02110958, \quad n02119022, \quad n02123394, \quad n02127052, \quad n02129165, \quad n02133161, \quad n02137549, \quad n02165456, \quad n02174001, \quad n02177972, \quad n02190166, \quad n02206856, \quad n02219486, \quad n02226429, \quad n02231487, \quad n02233338, \quad n02236044, \quad n02259212, \quad n02268443, \quad n02279972, \quad n02280649, \quad n02281787, \quad n02317335, \quad n02325366, \quad n02346627, \quad n02356798, \quad n02361337, \quad n02410509, \quad n02445715, \quad n02454379, \quad n02486410, \quad n02492035, \quad n02504458, \quad n02655020, \quad n02669723, \quad n02672831, \quad n02676566, \quad n02690373, \quad n02701002, \quad n02730930, \quad n02777292, \quad n02782093, \quad n02787622, \quad n02793495, \quad n02797295, \quad n02802426, \quad n02814860, \quad n02815834, \quad n02837789, \quad n02879718, \quad n02883205, \quad n02895154, \quad n02906734, \quad n02948072, \quad n02951358, \quad n02980441, \quad n02992211, \quad n02999410, \quad n03014705, \quad n03026506, \quad n03124043, \quad n03125729, \quad n03187595, \quad n03196217, \quad n03223299, \quad n03250847, \quad n03255030, \quad n03291819, \quad n03325584, \quad n03355925, \quad n03384352, \quad n03388043, \quad n03417042, \quad n03443371, \quad n03444034, \quad n03445924, \quad n03452741, \quad n03483316, \quad n03584829, \quad n03590841, \quad n03594945, \quad n03617480, \quad n03666591, \quad n03670208, \quad n03717622, \quad n03720891, \quad n03721384, \quad n03724870, \quad n03775071, \quad n03788195, \quad n03804744, \quad n03837869, \quad n03840681, \quad n03854065, \quad n03888257, \quad n03891332, \quad n03935335, \quad n03982430, \quad n04019541, \quad n04033901, \quad n04039381, \quad n04067472, \quad n04086273, \quad n04099969, \quad n04118538, \quad n04131690, \quad n04133789, \quad n04141076, \quad n04146614, \quad n04147183, \quad n04179913, \quad n04208210, \quad n04235860, \quad n04252077, \quad n04252225, \quad n04254120, \quad n04270147, \quad n04275548, \quad n04310018, \quad n04317175, \quad n04344873, \quad n04347754, \quad n04355338, \quad n04366367, \quad n04376876, \quad n04389033, \quad n04399382, \quad n04442312, \quad n04456115, \quad n04482393, \quad n04507155, \quad n04509417, \quad n04532670, \quad n04540053, \quad n04554684, \quad n04562935, \quad n04591713, \quad n04606251, \quad n07583066, \quad n07695742, \quad n07697313, \quad n07697537, \quad n07714990, \quad n07718472, \quad n07720875, \quad n07734744, \quad n07749582, \quad n07753592, \quad n07760859, \quad n07768694, \quad n07831146, \quad n09229709, \quad n09246464, \quad n09472597, \quad n09835506, \quad n11879895, \quad n12057211, \quad n12144580, \quad n12267677.

\section{\textsc{ImageNet-O} Classes}\label{app:classeso}

The 200 ImageNet classes that we selected for \textsc{ImageNet-O} are as follows.

\noindent`goldfish, Carassius auratus;' `triceratops;' `harvestman, daddy longlegs, Phalangium opilio;' `centipede;' `sulphur-crested cockatoo, Kakatoe galerita, Cacatua galerita;' `lorikeet;' `jellyfish;' `brain coral;' `chambered nautilus, pearly nautilus, nautilus;' `dugong, Dugong dugon;' `starfish, sea star;' `sea urchin;' `hog, pig, grunter, squealer, Sus scrofa;' `armadillo;' `rock beauty, Holocanthus tricolor;' `puffer, pufferfish, blowfish, globefish;' `abacus;' `accordion, piano accordion, squeeze box;' `apron;' `balance beam, beam;' `ballpoint, ballpoint pen, ballpen, Biro;' `Band Aid;' `banjo;' `barbershop;' `bath towel;' `bearskin, busby, shako;' `binoculars, field glasses, opera glasses;' `bolo tie, bolo, bola tie, bola;' `bottlecap;' `brassiere, bra, bandeau;' `broom;' `buckle;' `bulletproof vest;' `candle, taper, wax light;' `car mirror;' `chainlink fence;' `chain saw, chainsaw;' `chime, bell, gong;' `Christmas stocking;' `cinema, movie theater, movie theatre, movie house, picture palace;' `combination lock;' `corkscrew, bottle screw;' `crane;' `croquet ball;' `dam, dike, dyke;' `digital clock;' `dishrag, dishcloth;' `dogsled, dog sled, dog sleigh;' `doormat, welcome mat;' `drilling platform, offshore rig;' `electric fan, blower;' `envelope;' `espresso maker;' `face powder;' `feather boa, boa;' `fireboat;' `fire screen, fireguard;' `flute, transverse flute;' `folding chair;' `fountain;' `fountain pen;' `frying pan, frypan, skillet;' `golf ball;' `greenhouse, nursery, glasshouse;' `guillotine;' `hamper;' `hand blower, blow dryer, blow drier, hair dryer, hair drier;' `harmonica, mouth organ, harp, mouth harp;' `honeycomb;' `hourglass;' `iron, smoothing iron;' `jack-o'-lantern;' `jigsaw puzzle;' `joystick;' `lawn mower, mower;' `library;' `lighter, light, igniter, ignitor;' `lipstick, lip rouge;' `loupe, jeweler's loupe;' `magnetic compass;' `manhole cover;' `maraca;' `marimba, xylophone;' `mask;' `matchstick;' `maypole;' `maze, labyrinth;' `medicine chest, medicine cabinet;' `mortar;' `mosquito net;' `mousetrap;' `nail;' `neck brace;' `necklace;' `nipple;' `ocarina, sweet potato;' `oil filter;' `organ, pipe organ;' `oscilloscope, scope, cathode-ray oscilloscope, CRO;' `oxygen mask;' `paddlewheel, paddle wheel;' `panpipe, pandean pipe, syrinx;' `park bench;' `pencil sharpener;' `Petri dish;' `pick, plectrum, plectron;' `picket fence, paling;' `pill bottle;' `ping-pong ball;' `pinwheel;' `plate rack;' `plunger, plumber's helper;' `pool table, billiard table, snooker table;' `pot, flowerpot;' `power drill;' `prayer rug, prayer mat;' `prison, prison house;' `punching bag, punch bag, punching ball, punchball;' `quill, quill pen;' `radiator;' `reel;' `remote control, remote;' `rubber eraser, rubber, pencil eraser;' `rule, ruler;' `safe;' `safety pin;' `saltshaker, salt shaker;' `scale, weighing machine;' `screw;' `screwdriver;' `shoji;' `shopping cart;' `shower cap;' `shower curtain;' `ski;' `sleeping bag;' `slot, one-armed bandit;' `snowmobile;' `soap dispenser;' `solar dish, solar collector, solar furnace;' `space heater;' `spatula;' `spider web, spider's web;' `stove;' `strainer;' `stretcher;' `submarine, pigboat, sub, U-boat;' `swimming trunks, bathing trunks;' `swing;' `switch, electric switch, electrical switch;' `syringe;' `tennis ball;' `thatch, thatched roof;' `theater curtain, theatre curtain;' `thimble;' `throne;' `tile roof;' `toaster;' `tricycle, trike, velocipede;' `turnstile;' `umbrella;' `vending machine;' `waffle iron;' `washer, automatic washer, washing machine;' `water bottle;' `water tower;' `whistle;' `Windsor tie;' `wooden spoon;' `wool, woolen, woollen;' `crossword puzzle, crossword;' `traffic light, traffic signal, stoplight;' `ice lolly, lolly, lollipop, popsicle;' `bagel, beigel;' `pretzel;' `hotdog, hot dog, red hot;' `mashed potato;' `broccoli;' `cauliflower;' `zucchini, courgette;' `acorn squash;' `cucumber, cuke;' `bell pepper;' `Granny Smith;' `strawberry;' `orange;' `lemon;' `pineapple, ananas;' `banana;' `jackfruit, jak, jack;' `pomegranate;' `chocolate sauce, chocolate syrup;' `meat loaf, meatloaf;' `pizza, pizza pie;' `burrito;' `bubble;' `volcano;' `corn;' `acorn;' `hen-of-the-woods, hen of the woods, Polyporus frondosus, Grifola frondosa.'

Their WordNet IDs are as follows.

n01443537, \quad n01704323, \quad n01770081, \quad n01784675, \quad n01819313, \quad n01820546, \quad n01910747, \quad n01917289, \quad n01968897, \quad n02074367, \quad n02317335, \quad n02319095, \quad n02395406, \quad n02454379, \quad n02606052, \quad n02655020, \quad n02666196, \quad n02672831, \quad n02730930, \quad n02777292, \quad n02783161, \quad n02786058, \quad n02787622, \quad n02791270, \quad n02808304, \quad n02817516, \quad n02841315, \quad n02865351, \quad n02877765, \quad n02892767, \quad n02906734, \quad n02910353, \quad n02916936, \quad n02948072, \quad n02965783, \quad n03000134, \quad n03000684, \quad n03017168, \quad n03026506, \quad n03032252, \quad n03075370, \quad n03109150, \quad n03126707, \quad n03134739, \quad n03160309, \quad n03196217, \quad n03207743, \quad n03218198, \quad n03223299, \quad n03240683, \quad n03271574, \quad n03291819, \quad n03297495, \quad n03314780, \quad n03325584, \quad n03344393, \quad n03347037, \quad n03372029, \quad n03376595, \quad n03388043, \quad n03388183, \quad n03400231, \quad n03445777, \quad n03457902, \quad n03467068, \quad n03482405, \quad n03483316, \quad n03494278, \quad n03530642, \quad n03544143, \quad n03584829, \quad n03590841, \quad n03598930, \quad n03602883, \quad n03649909, \quad n03661043, \quad n03666591, \quad n03676483, \quad n03692522, \quad n03706229, \quad n03717622, \quad n03720891, \quad n03721384, \quad n03724870, \quad n03729826, \quad n03733131, \quad n03733281, \quad n03742115, \quad n03786901, \quad n03788365, \quad n03794056, \quad n03804744, \quad n03814639, \quad n03814906, \quad n03825788, \quad n03840681, \quad n03843555, \quad n03854065, \quad n03857828, \quad n03868863, \quad n03874293, \quad n03884397, \quad n03891251, \quad n03908714, \quad n03920288, \quad n03929660, \quad n03930313, \quad n03937543, \quad n03942813, \quad n03944341, \quad n03961711, \quad n03970156, \quad n03982430, \quad n03991062, \quad n03995372, \quad n03998194, \quad n04005630, \quad n04023962, \quad n04033901, \quad n04040759, \quad n04067472, \quad n04074963, \quad n04116512, \quad n04118776, \quad n04125021, \quad n04127249, \quad n04131690, \quad n04141975, \quad n04153751, \quad n04154565, \quad n04201297, \quad n04204347, \quad n04209133, \quad n04209239, \quad n04228054, \quad n04235860, \quad n04243546, \quad n04252077, \quad n04254120, \quad n04258138, \quad n04265275, \quad n04270147, \quad n04275548, \quad n04330267, \quad n04332243, \quad n04336792, \quad n04347754, \quad n04371430, \quad n04371774, \quad n04372370, \quad n04376876, \quad n04409515, \quad n04417672, \quad n04418357, \quad n04423845, \quad n04429376, \quad n04435653, \quad n04442312, \quad n04482393, \quad n04501370, \quad n04507155, \quad n04525305, \quad n04542943, \quad n04554684, \quad n04557648, \quad n04562935, \quad n04579432, \quad n04591157, \quad n04597913, \quad n04599235, \quad n06785654, \quad n06874185, \quad n07615774, \quad n07693725, \quad n07695742, \quad n07697537, \quad n07711569, \quad n07714990, \quad n07715103, \quad n07716358, \quad n07717410, \quad n07718472, \quad n07720875, \quad n07742313, \quad n07745940, \quad n07747607, \quad n07749582, \quad n07753275, \quad n07753592, \quad n07754684, \quad n07768694, \quad n07836838, \quad n07871810, \quad n07873807, \quad n07880968, \quad n09229709, \quad n09472597, \quad n12144580, \quad n12267677, \quad n13052670.

\end{document}